\documentclass{article}

\usepackage[preprint]{neurips_2026}

\usepackage[utf8]{inputenc}
\usepackage[T1]{fontenc}
\usepackage[hidelinks]{hyperref}
\usepackage{url}
\usepackage{booktabs}
\usepackage{amsfonts}
\usepackage{amsmath}
\usepackage{amssymb}
\usepackage{amsthm}
\usepackage{nicefrac}
\usepackage{microtype}
\usepackage{xcolor}
\usepackage{graphicx}
\usepackage[most]{tcolorbox}
\usepackage{enumitem}
\usepackage{needspace}
\usepackage{threeparttable}
\usepackage{subcaption}
\usepackage{tabularx}

\usepackage[ruled,noend]{algorithm2e}

\theoremstyle{plain}
\newtheorem{theorem}{Theorem}
\newtheorem{assumption}{Assumption}
\newtheorem{proposition}{Proposition}

\theoremstyle{remark}
\newtheorem{remark}{Remark}

\definecolor{accentgreen}{HTML}{2E7D5B}
\definecolor{bonebg}{HTML}{F5F2EA}

\tcolorboxenvironment{theorem}{
    enhanced, breakable,
    colback=accentgreen!4!bonebg!60!white,
    colframe=accentgreen,
    boxrule=0pt, leftrule=5pt,
    arc=0pt, outer arc=0pt,
    left=6pt, right=4pt, top=6pt, bottom=6pt,
    before skip=10pt, after skip=10pt
}
\tcolorboxenvironment{proposition}{
    enhanced, breakable,
    colback=accentgreen!4!bonebg!60!white,
    colframe=accentgreen,
    boxrule=0pt, leftrule=5pt,
    arc=0pt, outer arc=0pt,
    left=6pt, right=4pt, top=6pt, bottom=6pt,
    before skip=10pt, after skip=10pt
}

\tcolorboxenvironment{assumption}{
    enhanced, breakable,
    colback=accentgreen!4!bonebg!60!white,
    colframe=accentgreen,
    boxrule=0pt, leftrule=5pt,
    arc=0pt, outer arc=0pt,
    left=6pt, right=4pt, top=6pt, bottom=6pt,
    before skip=10pt, after skip=10pt
}

\newtcolorbox{divcondbox}{
    enhanced, breakable,
    colback=accentgreen!4!bonebg!60!white,
    colframe=accentgreen,
    boxrule=0pt, leftrule=5pt,
    arc=0pt, outer arc=0pt,
    left=6pt, right=4pt, top=2pt, bottom=2pt,
    before skip=8pt, after skip=8pt
}

\newcommand{\R}{\mathbb{R}}
\newcommand{\E}{\mathbb{E}}
\newcommand{\supp}{\mathrm{supp}}
\newcommand{\divg}{\nabla\cdot}
\DeclareMathOperator*{\argmin}{arg\,min}

\hypersetup{
  pdftitle={Beckmann Transport Models: From Autonomous Flows to One-Step Maps},
  pdfauthor={Lee Cheuk-Kit et al.}
}

\title{Beckmann Transport Models: \\
From Autonomous Flows to One-Step Maps}

\author{%
  Lee Cheuk-Kit$^{1}$ \quad
  Florentin Coeurdoux$^{2}$ \quad
  Yuyuan Chen$^{1}$ \quad
  Sophia Tang$^{3}$ \\[1mm]
  \textbf{Peter Potaptchik}$^{4}$ \quad
  \textbf{Yilun Du$^{1}$} \quad
  \textbf{Michael S. Albergo$^{1}$} \quad
  \textbf{Eric Vanden-Eijnden$^{2,5}$} \\[1mm]
  $^{1}$Harvard University \quad
  $^{2}$Capital Fund Management \quad
  $^{3}$University of Pennsylvania\\
  $^{4}$University of Oxford \quad
  $^{5}$New York University\\[1mm]
}
\begin{document}
\raggedbottom 

\maketitle
\addtocontents{toc}{\protect\setcounter{tocdepth}{-1}}

\begin{abstract}
We propose an instantiation of flow matching that relies on a time-independent velocity field---an autonomous flow---to exactly map between two distributions when the target is supported on a sufficiently lower-dimensional data manifold. We also show that the associated one-step generative map is constant in the direction of the autonomous velocity field and therefore satisfies a simple conservation equation, which motivates learning the map directly from samples. These autonomous flows and maps give a dynamical meaning to the flux constraint of Beckmann's transportation problem, and we show that, under suitable conditions, the construction extends beyond flow matching to other probability fluxes satisfying the same constraint. The resulting framework includes, for instance, a closed-form Coulomb transport closely related to Poisson Flow and a flow-matching-consistent version of Equilibrium Matching. We illustrate these distinctions and demonstrate the effectiveness of the autonomous flow and the one-step map on ImageNet $256\times256$.
\end{abstract}

\section{Introduction}
\label{sec:intro}

Standard flow matching \citep{lipman2022flow, albergo2023building, liu2022rectified} and diffusion models \citep{ho2020ddpm, song2021sgm} drive generation by transporting samples from a base distribution $\mu_0$ to a target distribution $\mu_1$ using a time-dependent velocity field (or \emph{drift}) learned by quadratic regression.

An intriguing alternative is to constrain the drift to be \emph{time-independent}: minimize the same regression loss over fields $b: \R^d \to \R^d$ that do not depend on $t$ and perform generation using the \emph{autonomous} equation $\dot X_t = b(X_t)$ solved with initial $X_0\sim \mu_0$. This is essentially the construction proposed by \citet{wang2025equilibrium} under the name \emph{Equilibrium Matching} (EqM), and it is appealing because it simplifies both the neural network architectures needed for generative models based on dynamical transport and the dynamical system itself. However, it raises an immediate question: \emph{does the autonomous flow with this time-independent drift actually transport $\mu_0$ to $\mu_1$?} The original EqM paper does not establish this transport property, and its proposed loss need not enforce the required source--sink balance.

We give a rigorous justification under explicit geometric and regularity conditions, and use it to derive new results. We show that when~$\mu_1$ is supported on a sufficiently regular manifold $M_1\subset\R^d$ of codimension at least two, the time-independent flow matching drift defines a valid transport: trajectories of the autonomous flow converge to $M_1$, and the resulting map pushes $\mu_0$ forward to $\mu_1$. The proof uses the finite FM current and the divergence equation satisfied by the drift. This structure also identifies a flow-matching-consistent loss for EqM, which differs from the loss originally proposed. The endpoint map $T$ is selected by following each autonomous trajectory to its endpoint and then freezing it there; it satisfies a simple conservation equation along its characteristics. This equation motivates a stop-gradient objective for training $T$ directly from samples; we validate the resulting one-step and iterated maps experimentally. At a structural level, these autonomous flows and maps give a \emph{dynamic} interpretation of admissible fluxes in Beckmann's transportation problem \citep{beckmann1952continuous, santambrogio2015ot}. For the FM current, the quadratic static functional is bounded above by the action of the original time-dependent flow. We refer to constructions of this kind collectively as \emph{Beckmann Transport Models} (BTM).

\begin{figure}
    \centering
    \includegraphics[width=1.0\linewidth]{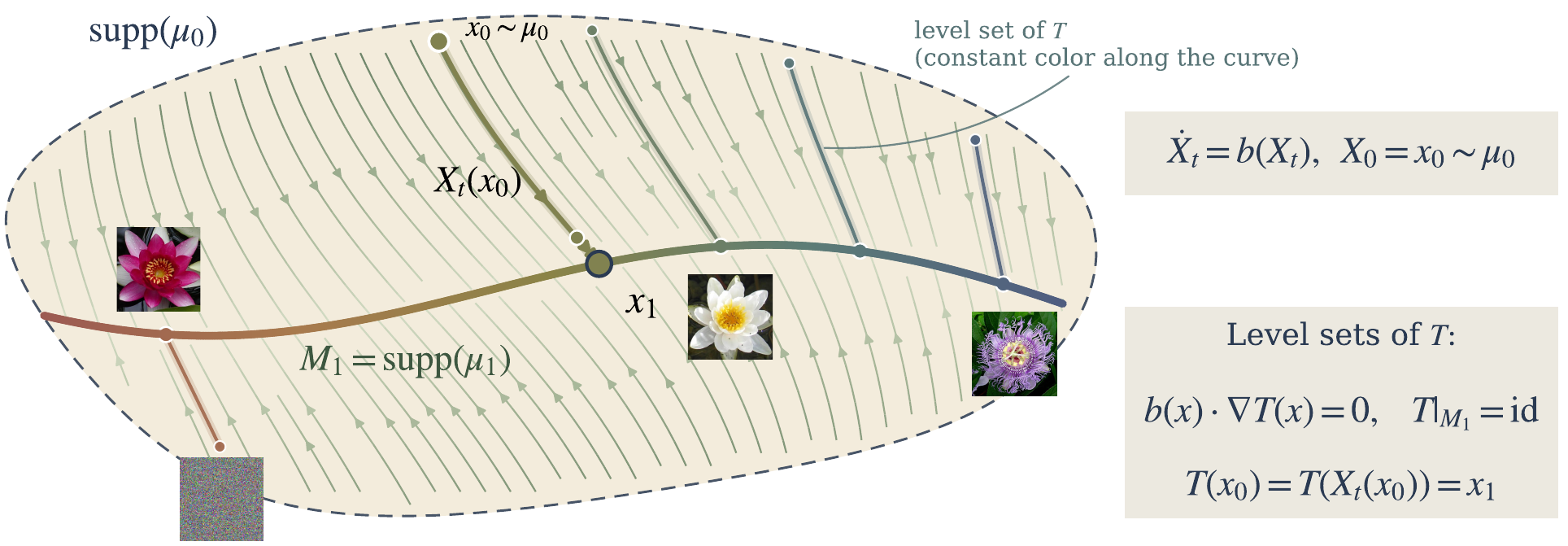}
    \caption{Overview of autonomous flow matching. Under Assumptions~\ref{ass:source_target}--\ref{ass:fm}, the autonomous flow $\dot X_t=b(X_t)$ transports $\mu_0$-almost every initial point to $M_1$ at hitting time $\tau$, defining the transport map $T=X_{\tau}$ that pushes $\mu_0$ forward to $\mu_1$. The endpoint construction selects $T$; its constancy along trajectories, $(d/dt)T(X_t)=0$, implies the conservation equation $b\cdot\nabla T=0$ and motivates a direct learning objective.}
    \label{fig:placeholder}
\end{figure}

Summarizing, our \textbf{main contributions} are:

\begin{itemize}[leftmargin=1em]
    \item We prove that, under the regularity and codimension conditions of Assumptions~\ref{ass:source_target}--\ref{ass:fm}, the time-independent FM drift defines a valid transport: the minimizer of the FM regression loss over time-independent fields $b$ generates a flow $\dot X_t=b(X_t)$ that pushes $\mu_0$ to $\mu_1$ (Proposition~\ref{prop:fmtransport}). This gives a flow-matching-consistent correction of Equilibrium Matching \citep{wang2025equilibrium}.
    \item More generally, for a finite positive weight $\nu$, we show that the unique minimizer of a weighted Beckmann current problem defines a valid autonomous transport when it points uniformly toward the target near $M_1$ (Theorem~\ref{thm:main}). Gradient-constrained FM provides a natural route to this minimizing current; the canonical Coulomb field with $\nu\equiv1$ is treated separately (Proposition~\ref{prop:coulomb}).
    \item We show that the map $x\mapsto X_t(x)$, which gives the position at time $t$ of the autonomous trajectory starting from $x$, satisfies the Eulerian equation $\partial_t X_t=b\cdot\nabla X_t$ (Proposition~\ref{prop:flow_eulerian}). The endpoint of each trajectory defines the one-step map $T(x)$, which pushes $\mu_0$ forward to $\mu_1$ (Proposition~\ref{prop:fmtransport}). This map is constant in the direction of the autonomous field $b$, so that $b\cdot\nabla T=0$, with $T=\mathrm{id}$ on $M_1$ (Theorem~\ref{thm:Tpde}).
    \item The conservation equation $b\cdot\nabla T=0$ motivates a stop-gradient objective for direct map learning. We prove that the population stop-gradient update vanishes at the true map $T$ (Theorem~\ref{thm:map_objective}) and evaluate the resulting one-step and iterated inference procedures experimentally.
\end{itemize}

We validate these results on 2D atomic targets and on the image benchmarks of \citet{wang2025equilibrium}, and show that the flow-matching-consistent autonomous field improves performance, and that the maps associated with these transports can be learned directly.

\subsection{Related work}
\label{sec:related}

\paragraph{Flow matching, diffusion, and equilibrium matching.}
Flow matching \citep{lipman2022flow, albergo2023building, albergo2023stochastic, liu2022rectified} and diffusion models \citep{ho2020ddpm, song2021sgm} construct differential equations with time-dependent drifts that transport between base and target distributions. The BTM drift minimizes the same FM loss, but over time-independent fields. BTM establishes an exact transport result for a flow-matching-consistent version of Equilibrium Matching \citep{wang2025equilibrium} and shows how the loss originally proposed there can produce a biased endpoint distribution. Equilibrium Forcing \citep{lillemark2026equilibriumforcing} empirically observes that EqM denoising trajectories need not reach the data manifold by the end of a fixed $[0,1]$ interval, and uses predicted noise levels to adapt integration and stopping. In BTM, sampling is governed by the first-hitting time, which could be learned directly (Appendix~\ref{app:hitting_time_learning}). Relatedly, \citet{christensen2025beyond} construct time-homogeneous denoising diffusions using Doob \(h\)-transforms with random horizons; for polar lower-dimensional data supports, generation terminates upon first hitting the support. Their construction is stochastic and score-based, rather than a deterministic transport derived from a Beckmann current. The choice $\nu=1$ also yields a closed-form Coulomb transport related to Poisson Flow \citep{xu2022pfgm}. In our construction the field is generated by the signed source--sink distribution $\mu_0-\mu_1$, whereas PFGM places the data charges on an augmented hyperplane and uses the induced source distribution on a distant hemisphere. PFGM++ \citep{xu2023pfgmpp} softens the Coulomb singularity through an additional dimension parameter, a different design choice from the weight $\nu$ in BTM.

\paragraph{Noise-unconditioned and blind generative models.}
\citet{kadkhodaie2021implicit} provide a blind-denoiser sampler, while \citet{sun2025noise} study removing noise conditioning. For linear FM, their population target matches our autonomous drift up to orientation, but their analysis bounds error relative to a conditioned sampler rather than proving an exact pushforward. \citet{kadkhodaie2026blind} give approximate guarantees under low intrinsic dimension, while \citet{sahraee2026geometry} study marginal-energy geometry and parameterization stability. Under stronger support and regularity assumptions, BTM instead proves exact endpoint transport at population level.

\paragraph{Relationship with optimal transport.}
The divergence condition is classical in OT \citep{santambrogio2015ot, villani2009optimal}. Standard flow matching realizes a transport without minimizing the Benamou--Brenier action \citep{benamou2000bb}; analogously, the FM instance of BTM gives a dynamical realization of a Beckmann flux without minimizing the associated cost, while Theorem~\ref{thm:main} treats the corresponding finite-mass weighted Beckmann minimizer \citep{beckmann1952continuous}. The variational form of the divergence equation also appears in NEIS \citep{cao2022neis, rotskoff2018neis} for partition-function estimation.

\paragraph{Consistency models, flow maps, and one-step generation.}
One-step or few-step generation has been pursued through consistency models \citep{song2023consistency}, rectified flow \citep{liu2022rectified}, Flow Map Matching \citep{boffi2024flowmapmatching, boffi2025consistencyflowmaps}, MeanFlow \citep{geng2025meanflow}, and Shortcut Models \citep{frans2024shortcut}. These methods start from an underlying time-dependent flow and distill or self-distill it into a one-step network. Drifting \citep{deng2026drifting} learns a one-step map directly, by evolving the network's pushforward during training, using an MMD distance to measure progress. BTM also learns the map directly, but exploits a different structural fact: the terminal map of the autonomous flow satisfies a stationary conservation equation along characteristics. This motivates a stop-gradient learning rule that requires no kernel; the present theory establishes population consistency of this update at the true map, while optimization and fixed-point selection are evaluated empirically.

\section{Autonomous transports and their associated maps}
\label{sec:fmtransport}

Standard flow matching \citep{lipman2022flow, albergo2023building, liu2022rectified} constructs a stochastic interpolant\footnote{We use $s$ for the interpolant time to avoid confusion with the time $t$ in the autonomous flow introduced later.}
\begin{equation}
\label{eq:interpolant}
I_s = \alpha_s x_0 + \beta_s x_1, \qquad x_0 \sim \mu_0,\ \ x_1 \sim \mu_1,\ \ s \in [0,1],
\end{equation}
where $\alpha_s$ and $\beta_s$ satisfy the boundary conditions $\alpha_0 = \beta_1 = 1$, $\alpha_1 = \beta_0 = 0$ so that the law of $I_s$ bridges between the base $\mu_0$ at time $s=0$ and the target $\mu_1$ at time $s=1$. The main result of the framework is that, at each fixed time $s<1$, the probability distribution $\mu_s$ of $I_s$ coincides with that of the solution to the probability flow ODE
\begin{equation}
    \label{eq:pfode:t}
    \dot Y_s = b_s(Y_s), \quad Y_0 \sim \mu_0,
\end{equation}
provided that the time-dependent drift is obtained by minimizing the standard FM loss over functions $\widehat b_s(x)$ that depend on $s$:
\begin{equation}
\label{eq:FMloss_td}
(b_s)_{s\in[0,1)}
=
\argmin_{(\widehat b_s)_{s\in[0,1)}}
\E_{s\sim U[0,1],x_0,x_1}
\bigl[|\widehat b_s(I_s)-\dot I_s|^2\bigr].
\end{equation}
Although the population minimization in~\eqref{eq:FMloss_td} separates over $s$, in practice one samples $s\sim U[0,1]$ and averages the loss over $s$, training a single time-conditioned neural network $\widehat b_s(x)$ that takes $(s,x)$ as input. At the population optimum, $b_s(x)=\E_{x_0,x_1}[\dot I_s\mid I_s=x]$ for each fixed $s$.

After learning $b_s$ from~\eqref{eq:FMloss_td}, inference integrates~\eqref{eq:pfode:t} from $Y_0\sim\mu_0$ to obtain samples $Y_1\sim\mu_1$. When $\mu_1$ is singular, $Y_1$ denotes the terminal limit $\lim_{s\uparrow1}Y_s$. Importantly, the pair $(\mu_s,b_s)$ satisfy the continuity equation
\begin{align}
\label{eq:ce}
    \partial_s \mu_s + \nabla \cdot (b_s \mu_s) = 0, \quad \mu_{s=0} = \mu_0,
\end{align}
with terminal law $\mu_{s=1}=\mu_1$.

\subsection{Autonomous flow matching}
Suppose instead that we minimize the \emph{same averaged FM loss} over time-independent functions $\widehat b(x)$, which do not take $s$ as input:
\begin{equation}
\label{eq:FMloss}
b
=
\argmin_{\widehat b:\R^d\to\R^d}
\E_{s \sim U[0,1],x_0,x_1}
\bigl[|\widehat b(I_s)-\dot I_s|^2\bigr],
\end{equation}
and consider the \emph{autonomous} ODE associated with this minimizer:
\begin{equation}
    \label{eq:ode}
    \dot X_t(x_0) = b(X_t(x_0)), \quad X_0(x_0) = x_0.
\end{equation}
The only difference between~\eqref{eq:FMloss_td} and~\eqref{eq:FMloss} is therefore the class of functions over which the same loss is minimized: $\widehat b_s(x)$ takes $s$ as an input, whereas $\widehat b(x)$ does not. Consequently, the population minimizer in~\eqref{eq:FMloss} is $b(x)=\E_{x_0,x_1,s}[\dot I_s\mid I_s=x]$, where the expectation also averages over $s\sim U[0,1]$. This gives a different drift, and hence a different flow, from~\eqref{eq:pfode:t}. A natural question is whether the autonomous ODE~\eqref{eq:ode} still transports $\mu_0$ to $\mu_1$. The answer is yes under the following assumptions.
\begin{assumption}[Regular source and singular target ($d\ge2$)]
\label{ass:source_target}
The base distribution $\mu_0(dx)=\rho_0(x)\,dx$ has a smooth, strictly positive, rapidly decaying density; in particular, $\rho_0$ has finite moments of every order. The target distribution $\mu_1$ is supported on a compact smooth embedded $k$-dimensional manifold $M_1\subset\R^d$ without boundary, where $k\le d-2$, and has a smooth, strictly positive density $\rho_1$ on $M_1$ with respect to its natural volume measure $\mathrm{vol}_{M_1}$.
\end{assumption}

The standard Gaussian base satisfies Assumption~\ref{ass:source_target}. Zero-padding a target in $\R^{d_1}$ into $\R^{d_1+d_2}$, as in \citep{xu2022pfgm}, increases the codimension by $d_2$; thus, if the padded target otherwise satisfies the compact-manifold and density conditions above, choosing $d_2\ge2$ gives $k\le d-2$. This addresses the codimension condition only. The atomic case $\mu_1=\sum_{j\in[N]}p_j\delta_{x_j}$ is included when $d\ge2$, with $k=0$, $M_1=\{x_j\}_{j\in[N]}$, and density $p_j$ at $x_j$. Precise regularity and measure conventions for Assumption~\ref{ass:source_target} are given at the beginning of Appendix~\ref{app:proof}.

\begin{assumption}[Straight FM interpolant with controlled endpoint behavior]
\label{ass:fm}
The interpolant has straight-line geometry,
\[
I_s=\alpha_s x_0+(1-\alpha_s)x_1,
\]
where $x_0\sim\mu_0$ and $x_1\sim\mu_1$ are independent, and $\alpha\in C^1([0,1])$, $\alpha_0=1$, $\alpha_1=0$, and $\dot\alpha_s<0$ for $0\le s<1$. The schedule satisfies
\[
-\dot\alpha_s=A(1-s)^\gamma(1+o(1))
\]
as $s\uparrow1$, for some $A>0$ and $\gamma\ge0$.
\end{assumption}

The restriction $\beta_s=1-\alpha_s$ gives a simple straight-path
construction for which one can prove directly that the time-integrated
flux carries no mass to infinity and points uniformly toward $M_1$
near the target: these are the two properties used to establish
Proposition~\ref{prop:fmtransport} below. The argument extends to more
general interpolants whenever their time-integrated flux is well
defined, sufficiently regular, and satisfies these two additional
geometric properties.
The divergence equation itself continues to follow from the endpoint
laws. Appendix~\ref{app:clock_geometry} explains these additional
checks.

Since $\dot I_s=\dot\alpha_s(x_0-x_1)$, $\gamma=0$ gives the interpolant a nonzero terminal speed whenever $x_0\ne x_1$, whereas $\gamma>0$ makes its speed vanish as $s\uparrow1$. For example, $\alpha_s=(1-s)^p$, with $p\ge1$, corresponds to $A=p$ and $\gamma=p-1$.

\begin{proposition}[Time-independent FM transport]
\label{prop:fmtransport}
Under Assumptions~\ref{ass:source_target}--\ref{ass:fm}, for $\mu_0$-almost every $x_0\in\R^d$, there exists a finite time $\tau(x_0)>0$ such that the solution of~\eqref{eq:ode} is defined and remains in $\R^d\setminus M_1$ for $0\le t<\tau(x_0)$, and
\[
T(x_0):=\lim_{t\uparrow\tau(x_0)}X_t(x_0)
\]
exists and belongs to $M_1$. In addition, the resulting endpoint map
$T$ satisfies $T_\sharp\mu_0=\mu_1$.
\end{proposition}

\emph{Proof sketch.}
Recall that the distribution $\mu_s$ of the interpolant satisfies the continuity equation~\eqref{eq:ce} with the time-dependent FM drift $b_s$. Integrating this equation from $s=0$ to $s=1$ gives $\divg j=\mu_0-\mu_1$, where
\[
j=\int_0^1 b_s\mu_s\,ds
\]
is the time-averaged probability flux. A direct calculation shows that the minimizer $b$ of the FM loss~\eqref{eq:FMloss} over time-independent fields satisfies $j=\nu b$, where $\nu=\int_0^1\mu_s\,ds>0$. Hence
\begin{divcondbox}
\refstepcounter{equation}\label{eq:divcond}%
\textbf{Divergence condition:}\qquad
$\divg(\nu b)=\mu_0-\mu_1$
\hfill(\theequation)
\end{divcondbox}
This equation identifies $\mu_0$ as the source and $\mu_1$ as the sink of the flux $j=\nu b$.

Since $\nu>0$, the fields $j$ and $b$ have the same oriented flow lines, differing only in their parametrization. Away from $M_1$, we have $\divg j=\rho_0>0$, so volumes transported by $j$ expand. Consequently, a positive mass of trajectories cannot remain trapped in a compact region away from the target. Moreover, the FM current carries no mass to infinity, while near $M_1$ it points toward the target. Hence a positive mass of trajectories can neither remain trapped away from $M_1$ nor escape to infinity, so almost every trajectory ends on $M_1$.

To understand the distribution of the endpoints, let $A\subseteq M_1$ and consider its basin $B_A=T^{-1}(A)$. Since $B_A$ is a union of flow lines, no flux crosses its lateral boundary. Flux balance therefore equates the source mass inside the basin with the sink mass at its endpoint:
\[
\mu_0(B_A)=\mu_1(A).
\]
Thus $T_\sharp\mu_0=\mu_1$. The full proof, including the rigorous justification of the endpoint and flux-balance arguments, is given in Appendix~\ref{app:proof}. \qed

\begin{remark}[Role of the singular target]
The role of the singular target is visible from the divergence equation~\eqref{eq:divcond}: it provides the localized sink at which the autonomous trajectories can end. Codimension also matters. In codimension one, a current satisfying the same divergence equation may pass through the target rather than stop there, and counterexamples show that the source--sink balance alone is not sufficient. For the straight FM construction in codimension at least two, the current instead points toward $M_1$ near the target, as proved in Appendix~\ref{app:proof}.
\end{remark}

\begin{remark}[The interpolant and autonomous clocks]
\label{rem:clock}
There are two distinct clocks. The schedule $\alpha_s$, with $s\in[0,1]$, governs the interpolant $I_s$ and the original time-dependent flow $\dot Y_s=b_s(Y_s)$. The autonomous flow~\eqref{eq:ode} instead runs on its own clock $t$. Its oriented streamlines, and hence the transport map $T$, depend only on the straight-line geometry $a x_0+(1-a)x_1$, not on the interpolant clock. In particular, $X_t(x_0)$ reaches $M_1$ at a trajectory-dependent autonomous time $\tau(x_0)$, generally different from the interpolant endpoint $s=1$. There is therefore no reason for $\mathrm{law}(X_t)$ to equal an intermediate interpolant law $\mu_s$; only the initial and terminal laws agree.

The behavior of the interpolant clock near $s=1$ nevertheless determines the magnitude of the autonomous drift near the target. Write $r(x)=\operatorname{dist}(x,M_1)$. Within the finite-horizon FM construction, the two endpoint regimes are
\[
\begin{array}{c|c|c}
\text{behavior of }\alpha_s\text{ at }s=1
& \text{autonomous drift near }M_1
& \text{hitting time}\\ \hline
\dot\alpha_1<0\;(\gamma=0)
& |b(x)|\asymp1
& \tau<\infty\\
\dot\alpha_1=0\;(\gamma>0)
& |b(x)|\asymp r(x)^{\gamma/(\gamma+1)}\to0
& \tau<\infty .
\end{array}
\]
Thus flattening the interpolant at $s=1$ changes the limiting magnitude of the autonomous drift, but not the finiteness of its hitting time. In the second regime, $\dot I_1=0$ and $b$ extends continuously by zero on $M_1$, so a threshold on $|b(X_t)|$ provides a practical stopping criterion. In the first regime, $b$ generally has no continuous zero extension, so this criterion is unavailable. The theoretical freezing convention is $X_t(x_0):=T(x_0)$ for $t\ge\tau(x_0)$.
\end{remark}

\subsection{Weighted Beckmann currents}

The divergence condition~\eqref{eq:divcond} does not by itself select a
unique current. For a fixed positive weight $\nu$ of finite mass,
consider the affine class
\begin{equation}
\label{eq:beckmann_class}
\mathcal J_\nu
:=
\left\{
j\in L^2(\nu^{-1}dx;\R^d):
\divg j=\mu_0-\mu_1
\right\},
\end{equation}
where the divergence equality is understood distributionally. Whenever
$\mathcal J_\nu$ is nonempty, it is a closed affine subset of the
Hilbert space $L^2(\nu^{-1}dx;\R^d)$, so the weighted Beckmann problem
\begin{equation}
\label{eq:beckmann_current}
j_\nu
:=
\argmin_{j\in\mathcal J_\nu}
\frac12\int_{\R^d}\frac{|j(x)|^2}{\nu(x)}\,dx
\end{equation}
has a unique solution. Set $b:=j_\nu/\nu$. The following result shows
that this current defines an autonomous transport when it points
uniformly toward the target near $M_1$. Write
$r(x):=\operatorname{dist}(x,M_1)$; because $M_1$ is a compact $C^3$
embedded manifold, $r$ is $C^2$ away from $M_1$ in a sufficiently small
tubular neighborhood.

\begin{theorem}[Finite-mass weighted Beckmann transport]
\label{thm:main}
Under Assumption~\ref{ass:source_target}, let $\nu\in C^1(\R^d\setminus M_1)$ be strictly positive and satisfy
\[
0<\int_{\R^d}\nu(x)\,dx<\infty.
\]
Assume that $\mathcal J_\nu$ is nonempty, let $j_\nu$ be the minimizer
in~\eqref{eq:beckmann_current}, and suppose that
$b:=j_\nu/\nu$ belongs to $C^1(\R^d\setminus M_1)$. Suppose moreover
that there exist $r_0,\kappa>0$ such that
\begin{equation}
\label{eq:terminal_beckmann}
j_\nu(x)\cdot\nabla r(x)\le-\kappa|j_\nu(x)|<0
\qquad (0<r(x)<r_0).
\end{equation}
Then, for $\mu_0$-almost every $x_0$, there exists $\tau(x_0)\in(0,\infty]$ such that the solution of~\eqref{eq:ode} is defined and remains in $\R^d\setminus M_1$ for $0\le t<\tau(x_0)$, and the unique limit
\begin{equation}
\label{eq:Tdef_general}
T(x_0):=\lim_{t\uparrow\tau(x_0)}X_t(x_0)
\end{equation}
belongs to $M_1$, where $t\uparrow\tau$ means $t\to\infty$ when
$\tau=\infty$. In addition, the resulting endpoint map $T$ satisfies
$T_\sharp\mu_0=\mu_1$.
\end{theorem}

\emph{Proof sketch.}
The divergence equation holds by the definition of $\mathcal J_\nu$.
Finite mass of $\nu$ and finite weighted energy imply
$j_\nu\in L^1$, so no positive amount of current can be lost at
infinity. Finally, \eqref{eq:terminal_beckmann} gives uniform
orientation toward $M_1$.
The general flux-balance theorem in Appendix~\ref{app:proof} then
shows that almost every trajectory has an endpoint on $M_1$ and that
the endpoint map pushes $\mu_0$ onto $\mu_1$. \qed

Condition~\eqref{eq:terminal_beckmann} controls the direction of the
trajectories but not their speed. If $\Gamma:[0,L]\to\R^d$ is the
arclength parametrization of the streamline from $x_0$ to $T(x_0)$,
then
\begin{equation}
\label{eq:hitting_integral}
\tau(x_0)=\int_0^L\frac{d\ell}{|b(\Gamma(\ell))|}.
\end{equation}
If, in addition, $|b(x)|\asymp r(x)^\sigma$ along the terminal part of
the trajectory, then condition~\eqref{eq:terminal_beckmann} gives
$-b\cdot\nabla r\asymp r^\sigma$, and the trajectory reaches $M_1$ in
finite time exactly when $\sigma<1$. In particular, when
$0<\sigma<1$, the drift vanishes at $M_1$, but the trajectory still
reaches it in finite time; extending $b$ by zero on $M_1$ then keeps it
there.

\begin{remark}[FM and Coulomb choices]
Let $j_{\mathrm{FM}}$ be the FM current of
Proposition~\ref{prop:fmtransport} and let
$\nu=\int_0^1\mu_s\,ds$ be its occupation weight. Then
$\int_{\R^d}\nu\,dx=1$, $j_{\mathrm{FM}}\in\mathcal J_\nu$, and
$j_{\mathrm{FM}}$ has finite weighted energy by
\eqref{eq:fm_action_bound}. If the autonomous FM regression is
restricted to gradient fields $b=\nabla\phi$, its population loss is,
up to an additive constant, equivalently the boundary functional
\begin{equation}
\label{eq:gradient_boundary_main}
\frac12\E\int_0^1|\nabla\phi(I_s)|^2\,ds
+\E[\phi(x_0)-\phi(x_1)]
=
\frac12\int_{\R^d}\nu|\nabla\phi|^2\,dx
+\int_{\R^d}\phi\,d(\mu_0-\mu_1).
\end{equation}
Indeed, the cross term is a pathwise derivative:
$\nabla\phi(I_s)\cdot\dot I_s=d\phi(I_s)/ds$. After the sign
change $V=-\phi$ and the identification $\mathcal D=\nu$,
\eqref{eq:gradient_boundary_main} is the same Dirichlet functional
used by \citet{cao2022neis} in their regular-density NEIS setting.
Appendix~\ref{app:gradient_fm} gives the equivalence, its sampling
form, and the regularity issues caused here by the singular target.
If this restricted problem admits a sufficiently regular minimizer,
its first-order condition gives
$\divg(\nu\nabla\phi)=\mu_0-\mu_1$. Since $\nu\nabla\phi$ is then
feasible and is orthogonal in $L^2(\nu^{-1}dx)$ to every finite-energy
divergence-free perturbation, it coincides with the unique minimizer
$j_\nu$. Thus gradient-constrained FM provides a way to
learn the minimizing current in~\eqref{eq:beckmann_current}, provided
its terminal orientation is verified. The gradient formulation also
has a particularly transparent dynamical interpretation. Along
$\dot X_t=\nabla\phi(X_t)$, the potential increases. Away from the
target, $\divg(\nu\nabla\phi)=\rho_0>0$ expands $\nu$-weighted
volumes, so a positive amount of source mass cannot remain in a compact
set disjoint from $M_1$; finite mass and energy exclude loss of current
at infinity. The remaining issue is terminal orientation. In
codimension one, the divergence equation fixes only the jump in normal
flux across $M_1$, not the sign of the flux on each side, so it does not
by itself prevent trajectories from passing through the target.
Appendix~\ref{app:gradient_fm} gives the precise argument and the
associated regularity conditions near $M_1$. We did not impose the
gradient constraint in our experiments: all reported FM models use
unconstrained vector fields, which are more expressive.

The finite-mass theorem does not include the Coulomb choice
$\nu\equiv1$, since both relevant finiteness conditions fail:
$\int_{\R^d}\nu\,dx=\infty$, and the behavior
$|j(x)|\asymp r(x)^{1-(d-k)}$ near $M_1$ makes
$\int_{\R^d}|j|^2\,dx$ diverge, logarithmically already in codimension two.
Nevertheless, it obeys the same transport conclusion because its
neutral far field gives no loss at infinity and its singular part points
uniformly toward $M_1$. This is stated next and proved in
Appendix~\ref{app:coulomb}.
\end{remark}

\begin{proposition}[Coulomb (Poisson) autonomous transport]
\label{prop:coulomb}
Under Assumption~\ref{ass:source_target}, let $\nu\equiv1$ and let $b=\nabla\phi$ be the canonical Coulomb field satisfying $\Delta\phi=\mu_0-\mu_1$, equivalently
\begin{equation}
\label{eq:coulomb_general}
b(x)=\frac1{\omega_d}\int_{\R^d}\frac{x-y}{|x-y|^d}\,(\mu_0-\mu_1)(dy),
\qquad x\notin M_1,
\end{equation}
where $\omega_d$ is the surface area of the unit sphere in $\R^d$. Then the first-hitting map of the autonomous flow $\dot X_t=b(X_t)$ transports $\mu_0$ to $\mu_1$, and its hitting time is finite for $\mu_0$-almost every initial point.
\end{proposition}

This is a training-free Coulomb transport closely related to the Poisson
Flow construction of \citet{xu2022pfgm}. Here the field is generated by
the signed distribution $\mu_0-\mu_1$, so that the prescribed base law is
an explicit source. Original PFGM instead places the data distribution on
an augmented hyperplane and obtains its source law from the flux on a
distant hemisphere. A direct extension of our transport theorem to such
singular source distributions would require a separate argument and is
outside the scope of this paper. The proof of
Proposition~\ref{prop:coulomb} combines the general flux-balance theorem
in Appendix~\ref{app:proof} with the Coulomb estimates given in
Appendix~\ref{app:coulomb}.

\subsection{Connection to Beckmann's transportation problem}
\label{sec:beckmann}

The classical Beckmann transportation problem
\citep{beckmann1952continuous,santambrogio2015ot} is
\begin{equation}
\label{eq:classical_beckmann}
\mathsf B(\mu_0,\mu_1)
:=
\inf_{\divg j=\mu_0-\mu_1}
\int_{\R^d}|j(x)|\,dx
=W_1(\mu_0,\mu_1).
\end{equation}
Here the infimum is over admissible currents and the integral denotes
their mass; precise conventions are given in Appendix~\ref{app:proof}.
The finite-mass currents constructed in this paper, in particular
$j_{\mathrm{FM}}$ and $j_\nu$, satisfy the same divergence constraint
and are therefore competitors in~\eqref{eq:classical_beckmann}, but we
do not claim that they solve this $L^1$ minimization problem.

Problem~\eqref{eq:beckmann_current} instead minimizes a quadratic
functional with the weight $\nu$ fixed. Its relation to the classical objective is exact
at the level of joint optimization: for every fixed current $j$,
\[
\inf_{\substack{\nu\ge0\\ \int_{\R^d}\nu\,dx=1}}
\int_{\R^d}\frac{|j|^2}{\nu}\,dx
=
\left(\int_{\R^d}|j|\,dx\right)^2,
\]
with equality for $\nu\propto|j|$. Thus optimizing both the weight and
the current would recover the square of the classical Beckmann problem;
this joint optimization is not performed here. For a fixed finite
weight, Theorem~\ref{thm:main} gives a dynamical realization of the
unique quadratic minimizer when its orientation near the target is
terminal. Proposition~\ref{prop:fmtransport} gives a separate
realization for the unconstrained FM current, and
Proposition~\ref{prop:coulomb} covers the infinite-mass Coulomb weight
$\nu\equiv1$. The structural analogy is:
\begin{center}
\renewcommand{\arraystretch}{1.25}
\begin{tabular}{@{}lll@{}}
\toprule
\textbf{Static formulation} & & \textbf{Dynamic / flow interpretation} \\
\midrule
Monge--Kantorovich optimal transport & $\longleftrightarrow$ & Benamou--Brenier (time-dependent flow) \\
Beckmann transportation             & $\longleftrightarrow$ & Beckmann Transport Models (autonomous flow) \\
\bottomrule
\end{tabular}
\end{center}
For the unconstrained FM current $j_{\mathrm{FM}}$ of
Proposition~\ref{prop:fmtransport}, with
$b_{\mathrm{FM}}=j_{\mathrm{FM}}/\nu$ and $\int_{\R^d}\nu\,dx=1$, feasibility,
Cauchy--Schwarz, and Jensen's inequality give
\begin{equation}
\label{eq:fm_action_bound}
W_1^2(\mu_0,\mu_1)
\le
\left(\int_{\R^d}|j_{\mathrm{FM}}|\,dx\right)^2
\le
\int_{\R^d}\frac{|j_{\mathrm{FM}}|^2}{\nu}\,dx
=\int_{\R^d} |b_{\mathrm{FM}}|^2 \nu\,dx
\;\le\; \int_0^1 \int_{\R^d} |b_s|^2\,d\mu_s\,ds.
\end{equation}
Thus the FM current provides an upper bound on the value of the
classical Beckmann problem, rather than solving it. In parallel, the
time-dependent FM path is an admissible competitor for the
Benamou--Brenier problem, and therefore
\[
W_2^2(\mu_0,\mu_1)
\le
\int_0^1 \int_{\R^d}|b_s|^2\,d\mu_s\,ds.
\]
This is the corresponding standard FM bound on the
Benamou--Brenier value. The two statements are analogous, but
$W_2^2(\mu_0,\mu_1)$ need not be bounded by the static quadratic
functional in the middle of~\eqref{eq:fm_action_bound}, because $\nu$
is not the occupation density of the autonomous flow. The proof of
\eqref{eq:fm_action_bound} is in Appendix~\ref{app:cost_bound}.

\subsection{The flow map}
\label{sec:pde_viewpoint}
 
Consider any of the autonomous drifts in
Proposition~\ref{prop:fmtransport}, Theorem~\ref{thm:main}, or
Proposition~\ref{prop:coulomb}. When $\tau(x_0)<\infty$, we extend
$X_t(x_0)$ to all later times by \emph{freezing the flow after arrival},
setting $X_t(x_0):=T(x_0)$ for $t\ge\tau(x_0)$; when
$\tau(x_0)=\infty$, no extension is needed. Under this convention,
$X_t$ inherits the semigroup property of the autonomous ODE,
\begin{equation}
\label{eq:semigroup}
X_{t+s} = X_t \circ X_s \qquad \text{for all } s, t \ge 0,
\end{equation}
with $X_0 = \mathrm{id}$, since $X_t \circ X_s$ leaves $M_1$ fixed once any trajectory has reached it. We now show that $X_t$ also satisfies a transport equation in $t$.

\begin{proposition}[Eulerian form of the flow]
\label{prop:flow_eulerian}
Let $b$ be one of the autonomous drifts in
Proposition~\ref{prop:fmtransport}, Theorem~\ref{thm:main}, or
Proposition~\ref{prop:coulomb}, and let $X_t$ be the corresponding
autonomous flow~\eqref{eq:ode}, with the freezing convention above.
Viewed as a flow map of the initial condition $x\in\R^d$, $X_t(x)$
satisfies, wherever the spatial derivative exists and for almost every
$t$, the \emph{Eulerian} equation
\begin{equation}
\label{eq:eulerian}
\partial_t X_t(x) = b(x) \cdot \nabla X_t(x), \quad X_0(x) = x, \quad \text{for } x \notin M_1, \qquad X_t(x) = x \quad \text{for } x \in M_1.
\end{equation}
Here $b\cdot\nabla X_t:=DX_t\,b$ is the Jacobian--vector product,
equivalently $(b\cdot\nabla X_t)^i=b\cdot\nabla X_t^i$ componentwise.
\end{proposition}

\begin{proof}
The semigroup identity \eqref{eq:semigroup} gives $X_{t+s}(x)=X_t(X_s(x))$. At points where the relevant derivatives exist, differentiating in $s$ at $s=0$ yields
\[
\partial_tX_t(x)=b(x)\cdot\nabla X_t(x).
\]
For $t<\tau(x)$ this is the usual autonomous flow identity. For $t>\tau(x)$, the left-hand side vanishes by freezing and the same semigroup argument gives $b\cdot\nabla X_t=0$. At the hitting instant the time derivative need not exist, so the equation is understood almost everywhere in $t$. The boundary condition $X_t(x)=x$ for $x\in M_1$ is the freezing convention.
\end{proof}

\subsection{The transport map}
Proposition~\ref{prop:fmtransport}, Theorem~\ref{thm:main}, and
Proposition~\ref{prop:coulomb} define endpoint maps associated with
their autonomous flows. With the freezing convention above, each can
equivalently be written
\[
T(x)=\lim_{t\to\infty}X_t(x).
\]
Thus $T$ is uniquely selected as the terminal limit of the flow, rather
than as an arbitrary solution of a stationary boundary-value problem.
When $\tau(x)<\infty$, one may evaluate $T$ by integrating the ODE until
first hitting; when $\tau(x)=\infty$, it is obtained as the terminal
limit. The following conservation law motivates learning the terminal
map directly.

\begin{theorem}[Conservation equation along characteristics]
\label{thm:Tpde}
Let $T$ be the endpoint map of Proposition~\ref{prop:fmtransport}, Theorem~\ref{thm:main}, or Proposition~\ref{prop:coulomb}. The endpoint is constant along $\mu_0$-almost every characteristic:
\[
T(X_t(x))=T(x),
\qquad 0\le t<\tau(x).
\]
After setting $T(x)=x$ on $M_1$, this gives, wherever $T$ is differentiable,
\begin{equation}
\label{eq:T_pde}
b(x) \cdot \nabla T(x) = 0 \quad \text{for } x \notin M_1, \qquad T(x) = x \quad \text{for } x \in M_1.
\end{equation}
\end{theorem}

The proof is in Appendix~\ref{app:proof}. The terminal-limit construction uniquely defines $T$ up to a $\mu_0$-null set. By contrast, if \eqref{eq:T_pde} is considered as a standalone stationary boundary-value problem, the boundary value must be understood as a trace along terminal characteristics; pointwise values on the lower-dimensional set $M_1$ alone do not ensure uniqueness.

Theorem~\ref{thm:main} selects a finite-mass weighted Beckmann current and supplies sufficient terminal conditions for its endpoint map, while Theorem~\ref{thm:Tpde} records the conservation law obeyed by that map. Propositions~\ref{prop:fmtransport} and~\ref{prop:coulomb} provide the corresponding unconstrained-FM and Coulomb constructions. Global continuity is neither asserted nor generally possible: for example, a nonconstant map from connected $\R^d$ to a finite atomic target must be discontinuous across basin boundaries.
 
\subsection{Direct learning of the transport map}
\label{sec:direct:l}
From the conservation equation \eqref{eq:T_pde} we could in principle learn $T$ by minimizing the squared residual $\E[|b \cdot \nabla T|^2]$. This requires access to the drift $b$, which the FM regression provides. A residual loss, however, does not prescribe how a partially trained map should relate to the flow-map evolution. Proposition~\ref{prop:flow_eulerian} provides a more direct motivation: the transport map is the terminal limit of the flow-map evolution, and a formal explicit-Euler step of \eqref{eq:eulerian} reads
\begin{equation}
\label{eq:euler_step}
X^{(k+1)}(x) = X^{(k)}(x) + \eta\, b(x) \cdot \nabla X^{(k)}(x).
\end{equation}
Motivated by this update, the algorithm tested in this paper uses a single network on both sides and stops gradients through the right-hand side. This is similar in spirit to the Drifting framework of \citet{deng2026drifting} but avoids a kernel; see Appendix~\ref{app:drifting} for a comparison. Because the same network is used on both sides, this optimization is not an exact fitted Euler iteration, and its parameter-space trajectory is not asserted to coincide with the flow-map evolution. The factor $\eta$ only rescales the stopped transport gradient relative to the learning rate and boundary term, so we absorb it into those optimization parameters below.
 
In the FM setup we do not need $b$ explicitly: the regression identity $b(x) = \E_{s, x_0, x_1}[\dot I_s \mid I_s = x]$ allows us to use the conditional sample $\dot I_s \cdot \nabla T(I_s)$. Adding a soft penalty for the boundary value in \eqref{eq:T_pde} yields the tested population objective.

\begin{theorem}[Consistency of the stop-gradient update]
\label{thm:map_objective}
Under Assumptions~\ref{ass:source_target}--\ref{ass:fm}, suppose that the terminal map $T$ is differentiable almost everywhere with respect to the FM occupation measure and that the expectations below are finite. Then the population stop-gradient update vanishes at $T$ for the objective
\begin{equation}
\label{eq:Tloss}
\mathcal{L}(T) = \E_{s,x_0,x_1}\bigl[\bigl|T(I_s) - \mathrm{sg}\bigl(T(I_s) + \dot I_s \cdot \nabla T(I_s)\bigr)\bigr|^2\bigr] + \lambda\,\E_{x_1}[|T(x_1) - x_1|^2].
\end{equation}
\end{theorem}
\begin{proof}
Let $H$ be an admissible test variation. Because gradients are stopped through the target, the variation used by the stop-gradient update is
\[
-2\,\E\bigl[(\dot I_s\cdot\nabla T(I_s))\cdot H(I_s)\bigr]
=-2\int_{\R^d}\nu(x)\,(b(x)\cdot\nabla T(x))\cdot H(x)\,dx=0,
\]
where the conditional-expectation identity for $b$ and Theorem~\ref{thm:Tpde} were used. The corresponding variation of the boundary term vanishes because $T(x_1)=x_1$ for $\mu_1$-almost every $x_1$.
\end{proof}

The theorem establishes that the proposed learning rule is consistent with the true terminal map at population level. As is typical for stop-gradient objectives, it does not by itself prove uniqueness of the fixed point or convergence of neural-network optimization. The terminal solution is selected theoretically by the frozen-flow limit, while the practical optimization behavior is evaluated empirically.

\begin{figure}[h]
    \centering
    \includegraphics[width=1.0\linewidth]{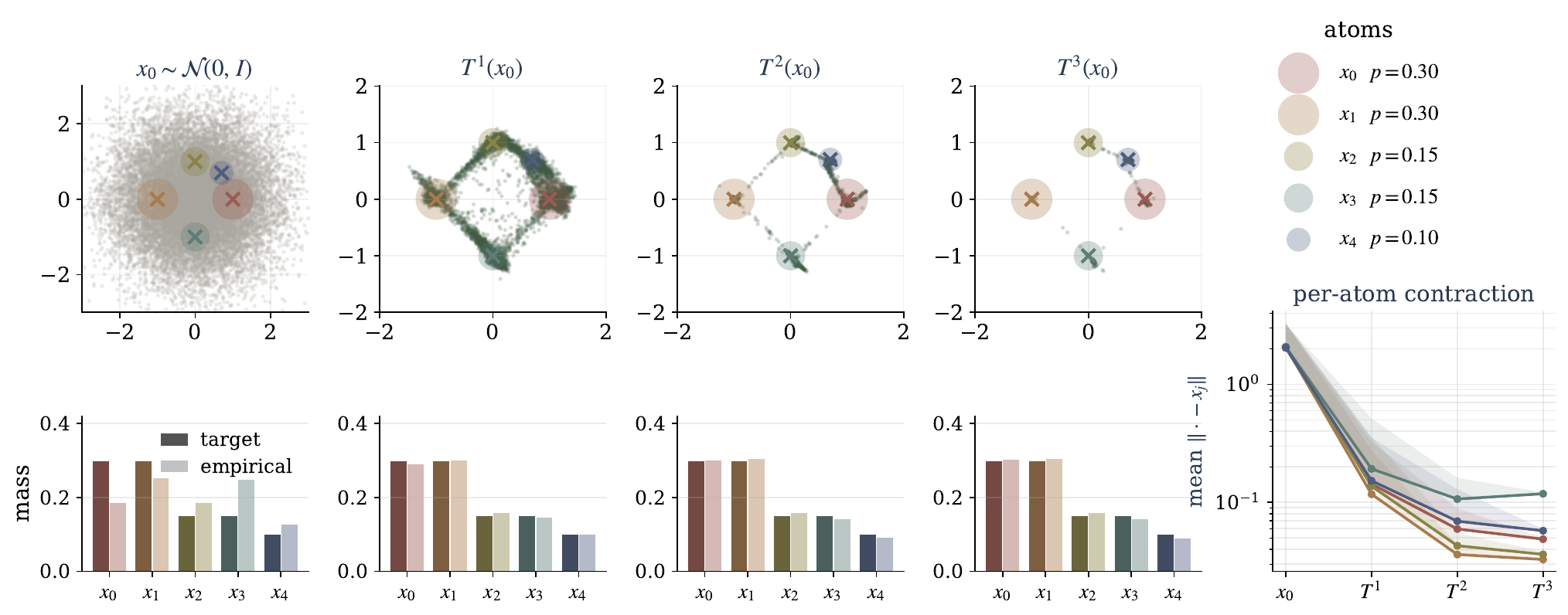}
    \caption{Empirical effect of iterating a map learned with \eqref{eq:Tloss} on a 5-mode atomic target. Repeated application sharpens the clusters and moves samples closer to the singular support.}
    \label{fig:cluster}
\end{figure}

\paragraph{Iterated application of the learned map.}
Once the map is trained, sampling at inference is a single forward pass: $x_1=T_\theta(x_0)$ for $x_0\sim\mu_0$. The semigroup property of the exact flow motivates testing whether repeated application of a partially trained network improves its approximation to the terminal map. Although a partially trained network is not guaranteed to equal $X_t$ for any $t$, empirically its iterates sharpen samples, as illustrated in Figure~\ref{fig:cluster}.

\subsection{Equilibrium Matching as Approximate Self-Stopping BTM}

\citet{wang2025equilibrium} proposed Equilibrium Matching via a loss that resembles \eqref{eq:FMloss} but pairs the linear interpolant $I_s = (1-s)x_0 + sx_1$ with a target $c_s(x_1 - x_0)$ scaled by a non-trivial schedule:
\begin{equation}
\label{eq:EMloss}
\E_{s,x_0,x_1}\bigl[|b((1-s)x_0 + sx_1) - c_s(x_1 - x_0)|^2\bigr].
\end{equation}

The schedule is chosen with the aim of making $b$ vanish as the flow approaches the support $\supp(\mu_1)$. Sampling is then done by following the drift over an indefinite amount of time.

This coincides with the loss in \eqref{eq:FMloss} only when $c_s \equiv 1$, since $\dot I_s=x_1-x_0$ for the stated linear interpolant. For other choices of $c_s$, the regression target is not the velocity of that interpolant and therefore does not enforce the divergence condition~\eqref{eq:divcond}. It will generally introduce bias and fail to transport $\mu_0$ to $\mu_1$. In the examples of Section~\ref{sec:experiments}, the flow defined by $\dot{\widetilde X}_t(x_0)=\widetilde b(\widetilde X_t(x_0))$ still converges to $M_1$, but with incorrect endpoint weights.

The resolution, which is what \eqref{eq:FMloss} implements, is to treat the schedule as part of the interpolant rather than as a separate weight on the target. To satisfy the desired tail behaviour, one may choose the straight interpolant of Assumption~\ref{ass:fm} with $\dot\alpha_1=0$, equivalently $\dot I_1=0$. The resulting autonomous drift extends continuously by zero to the support of $\mu_1$. Our experiment in Section~\ref{sec:experiments} explores one such choice and confirms that this correction improves the quality reported by \citet{wang2025equilibrium} at the same training cost.

\begin{algorithm}[H]
\caption{Learning the autonomous drift~$b$}\label{alg:learn_b}
\KwIn{Samples $\{x_0^{(n)}\} \sim \mu_0$, $\{x_1^{(m)}\} \sim \mu_1$; straight interpolant with $\beta_s=1-\alpha_s$; network $b_\theta$; learning rate $\eta$}
\For{$\mathrm{iteration} = 1, 2, \ldots$}{
    Sample a minibatch $\mathcal{B}$\;
    \For{$k \in \mathcal{B}$}{
        Sample $s_k \sim U([0,1])$, $x_0^k \sim \mu_0$, $x_1^k \sim \mu_1$; set $I_k = \alpha_{s_k} x_0^k + \beta_{s_k} x_1^k$, $\dot I_k = \dot\alpha_{s_k} x_0^k + \dot\beta_{s_k} x_1^k$\;
    }
    $\theta \leftarrow \theta - \eta\, \nabla_\theta L$, \ where \ $\displaystyle L = \frac{1}{|\mathcal{B}|} \sum_{k \in \mathcal{B}} \bigl|b_\theta(I_k) - \dot I_k\bigr|^2$\;
}
\Return $b_\theta$
\end{algorithm}

\begin{algorithm}[H]
\caption{Learning the transport map~$T$ with the base objective}\label{alg:learn_T}
\KwIn{Samples $\{x_0^{(n)}\} \sim \mu_0$, $\{x_1^{(m)}\} \sim \mu_1$; straight interpolant with $\beta_s=1-\alpha_s$; network $T_\theta$; boundary weight $\lambda$; learning rate $\eta$}
\For{$\mathrm{iteration} = 1, 2, \ldots$}{
    Sample a minibatch $\mathcal{B}$\;
    \For{$k \in \mathcal{B}$}{
        Sample $s_k \sim U([0,1])$, $x_0^k \sim \mu_0$, $x_1^k \sim \mu_1$;
        set $I_k = \alpha_{s_k} x_0^k + \beta_{s_k} x_1^k$, $\dot I_k = \dot\alpha_{s_k} x_0^k + \dot\beta_{s_k} x_1^k$,
        $\widehat T_k = T_\theta(I_k) + \dot I_k \cdot \nabla T_\theta(I_k)$\;
    }
    $\theta \leftarrow \theta - \eta\, \nabla_\theta L$, \ where \ $\displaystyle L = \frac{1}{|\mathcal{B}|} \sum_{k \in \mathcal{B}} \Bigl[\bigl|T_\theta(I_k) - \mathrm{sg}(\widehat T_k)\bigr|^2 + \lambda\,\bigl|T_\theta(x_1^k) - x_1^k\bigr|^2\Bigr]$\;
}
\Return $T_\theta$
\end{algorithm}

\section{Experiments}
\label{sec:experiments}
We validate several claims experimentally: (i)~the original EqM loss~\eqref{eq:EMloss} produces biased pushforward weights, the geometry of which is directly visible in the basin structure of the learned drift, and the consistent interpolant~\eqref{eq:FMloss} corrects this bias at no additional cost; (ii) on lower-dimensional targets, the learned map $T_\theta$ allows one-step generation and improves under iteration; and (iii) BTM scales to large-scale image generation, matching and modestly improving the quality reported by~\citet{wang2025equilibrium}.

In Appendix~\ref{app:experiments}, we further study the dependence of the bias on the schedule exponent, the trade-off between convergence speed and correctness controlled by a tail parameter, and the training-free Coulomb transport of Appendix~\ref{app:coulomb}. We defer a larger-scale evaluation of the direct map learning algorithm to future studies and provide preliminary results in Appendix~\ref{app:prelim-image}.

\begin{figure}[h!]
  \centering
  \includegraphics[width=1.0\textwidth]{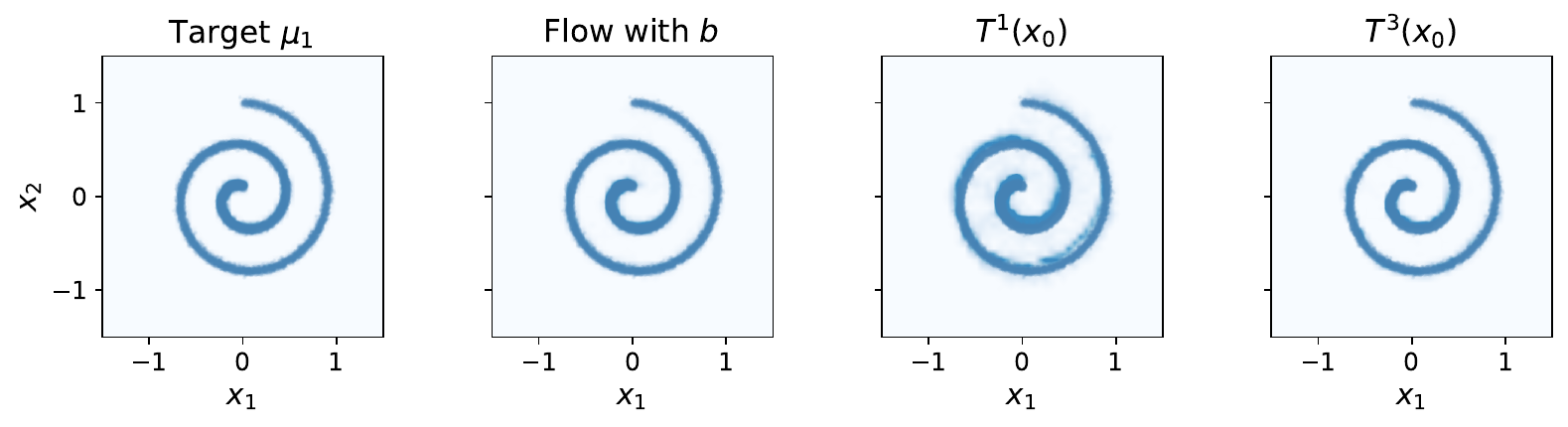}
  \caption{\textbf{Flow vs.\ map generation.} Spiral target embedded in $d = 6$ with $\mu_0 = \mathcal{N}(0, I_6)$. From left: target $\mu_1$; autonomous flow $\dot X_t = b(X_t)$; one application of $T_\theta$ (1-NFE); three applications of $T_\theta$ (3-NFE). Iteration empirically sharpens a partially trained map; after fuller training, one application matches the flow visually.}
  \label{fig:spiral_iterated}
  \label{fig:one_step}
\end{figure}

\paragraph{BTM corrects bias in Equilibrium Matching.} We test whether BTM, learned with the consistent loss~\eqref{eq:FMloss}, corrects the bias produced by the original EqM loss~\eqref{eq:EMloss}. We consider a two-dimensional example where $\mu_0 = \mathcal{N}(0, I_2)$ and $\mu_1 = \sum_{j=1}^5 p_j\,\delta_{x_j}$, with deliberately unequal weights $p = (0.30,\, 0.30,\, 0.15,\, 0.15,\, 0.10)$ chosen so that weight errors are immediately visible. For BTM, the five basin areas match the target weights to within measurement error ($\mathrm{MAE} = 0.005$). For EqM, the two largest basins expand while the smallest nearly disappears ($\mathrm{MAE} = 0.102$, a $20\times$ error). The mechanism is visible in the separatrices: the EqM drift satisfies $b(x_j) \approx 0$ because $c_s \to 0$ forces the regression target to vanish at $s = 1$, but this does not enforce the divergence condition~\eqref{eq:divcond}, so the flow reaches $M_1$ with the wrong weights. The consistent interpolant ties the regression target to $\dot I_s$, encoding the source--sink balance $\mu_0 - \mu_1$; the minimizer then satisfies~\eqref{eq:divcond} by construction.
\begin{figure}
  \centering
  \includegraphics[width=0.9\textwidth]{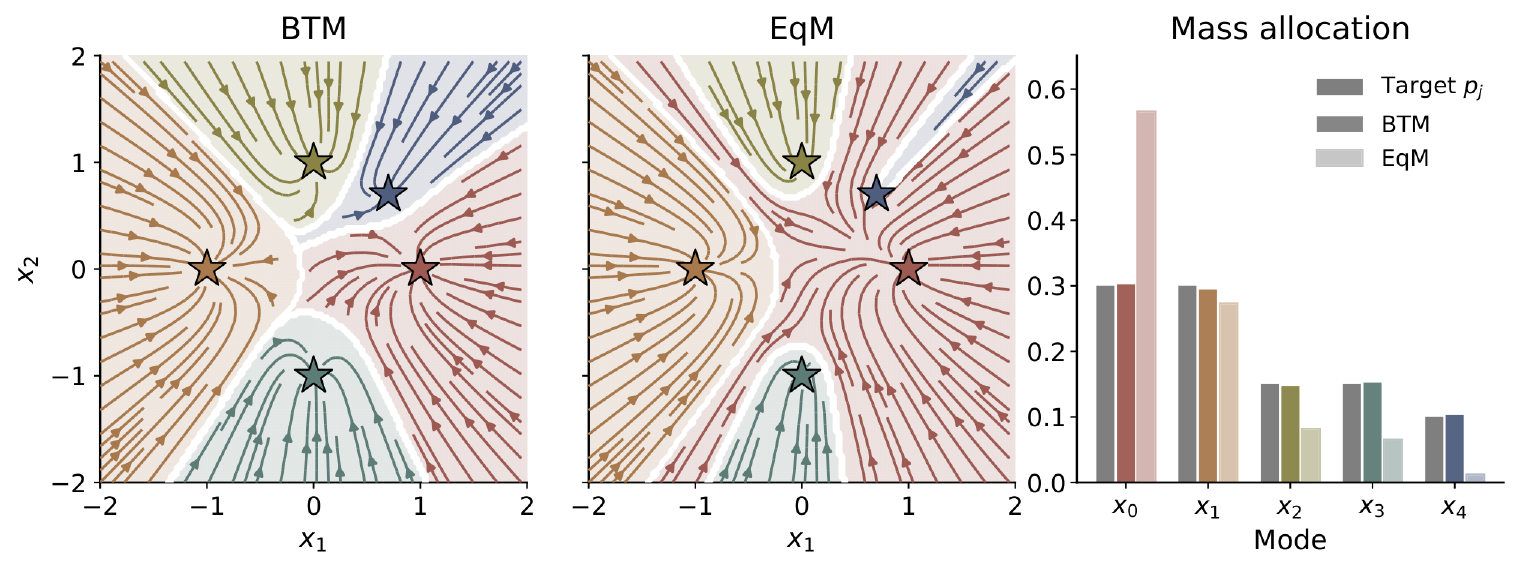}
  \caption{\textbf{Basins of attraction} for BTM (left) and EqM (right), with schedule $c_s = (1-s)^{0.8}$ and $J = 5$ atoms. Each color identifies the set of initial conditions that converge to the corresponding atom; white lines are the separatrices. The EqM loss does not enforce the source--sink balance and produces biased endpoint weights, which the consistent BTM loss corrects without additional computational cost.}
  \label{fig:weighted_vs_unweighted}
\end{figure}
\paragraph{Flow vs.\ map generation (Figure~\ref{fig:spiral_iterated}).}
We use the two-dimensional spiral distribution embedded in $d = 6$ dimensions ($\mu_0 = \mathcal{N}(0, I_6)$) using $I_s=(1-s) x_0 + s x_1$ to compare generation with the autonomous flow associated with $b$ against the learned one-step map $T_\theta$ and its iterates. The autonomous flow recovers the target sharply, while a one-shot map from a partially trained network is visibly diffuse around the spiral. Applying the trained network three times sharpens generation substantially and brings it close to the flow; after fuller training, one application matches the flow visually. This supports the empirical use of iteration as a way to trade additional forward passes for sample refinement. It does not require, or establish, that an intermediate network equals an exact flow map $X_t$. Crucially, inference here uses one or three network evaluations rather than numerical ODE integration.
\paragraph{BTM correction improves EqM image generation without additional cost (Figure~\ref{fig:curated_samples}).}
We train the consistent loss~\eqref{eq:FMloss} on class-conditional ImageNet $256 \times 256$, using the XL/2 architecture and the same training budget as~\citet{wang2025equilibrium}, with a self-stopping interpolant detailed in Appendix~\ref{app:self-stopping-interpolant}, with the same sampling scheme as EqM. The corrected model (EqM-XL/2) achieves FID $= \mathbf{1.87}$ versus FID $= 1.90$ for the uncorrected EqM-XL/2 under the same conditions. The gain is modest, as expected if the bias is primarily a weight-allocation error and the class distribution is nearly uniform. It requires no change in architecture, training cost, or inference procedure. Curated samples are shown in Figure~\ref{fig:curated_samples}.

\begin{minipage}[t]{0.55\textwidth}
  \vspace{0pt} 
  \paragraph{BTM enables one-step image generation.} 
  We train a one-step map using the stop-gradient loss~\eqref{eq:Tloss} with an XL/2 architecture on an SD-VAE latent space, omitting time embeddings. For these ImageNet map experiments only, we additionally use adaptive weighting, gradient balancing, and a slightly noisy boundary anchor; these empirical modifications are detailed in Appendix~\ref{app:map-expt} and are not part of Theorem~\ref{thm:map_objective}. As shown in Table~\ref{tab:fid_comparison}, under the same budget as~\citet{wang2025equilibrium}, BTM achieves an FID of 17.58. We compare with iCT \citep{song2024improved}, Shortcut Models \citep{frans2024shortcut}, SiT \citep{ma2024sitexploringflowdiffusionbased}, and MeanFlow \citep{geng2025meanflow}. While our unguided one-step result remains modest, the strongest competing one-step methods rely heavily on classifier-free guidance (CFG) for sample quality. Extending CFG to BTM remains an open direction for future work.
\end{minipage}\hfill
\begin{minipage}[t]{0.41\textwidth}
  \vspace{0pt} 
  \centering
  \footnotesize
  \setlength{\tabcolsep}{3.5pt}
  \renewcommand{\arraystretch}{1.1}
  \begin{tabular}{lllc}
    \toprule
    \textbf{Model} & \textbf{NFE} & \textbf{Guidance} & \textbf{FID} $\downarrow$ \\
    \midrule
    iCT & 1 & None & 30.10 \\
    \textbf{BTM (Ours)} & \textbf{1} & \textbf{None} & \textbf{17.58} \\
    Shortcut Models & 1 & CFG & 10.60 \\
    SiT (w/o CFG) & 250 & None & 8.30 \\
    MeanFlow & 1 & CFG & 3.43 \\
    SiT (w/ CFG) & $2\times250$ & CFG & 2.06 \\
    \bottomrule
  \end{tabular}
  \captionof{table}{FID comparison against recent baselines. All numbers use the same evaluation protocol.}
  \label{tab:fid_comparison}
\end{minipage}

\begin{figure}[t]
  \centering
  \begin{minipage}[c]{0.62\linewidth}
    \centering
    \includegraphics[width=\linewidth]{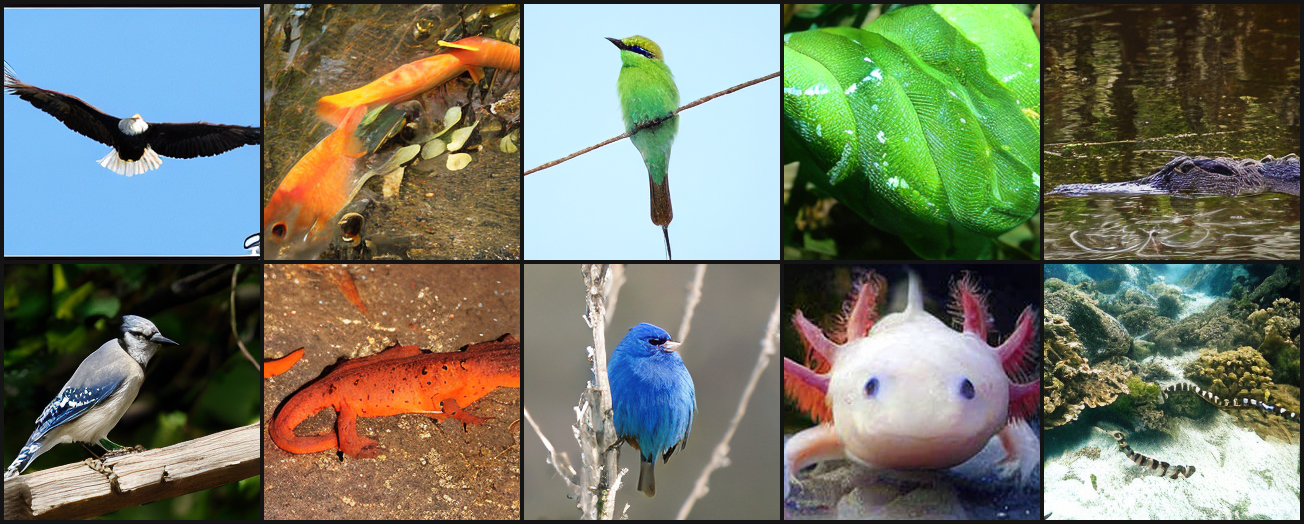}
  \end{minipage}\hfill
  \begin{minipage}[c]{0.36\linewidth}
    \centering
    \footnotesize
    \setlength{\tabcolsep}{4pt}
    \renewcommand{\arraystretch}{1.1}
    \begin{tabular}{llc}
      \toprule
      \textbf{Model} & \textbf{Method} & \textbf{FID} $\downarrow$ \\
      \midrule
      StyleGAN-XL & GAN       & 2.30 \\
      VDM++       & Diffusion & 2.12 \\
      DiT-XL/2    & Diffusion & 2.27 \\
      SiT-XL/2    & FM        & 2.06 \\
      \midrule
      EqM-XL/2    & EqM       & 1.90 \\
      EqM-XL/2    & BTM       & \textbf{1.87} \\
      \bottomrule
    \end{tabular}
  \end{minipage}
  \caption{\textbf{Left:} Curated class-conditional samples from the EqM-XL/2 architecture on ImageNet $256\times256$. Each image corresponds to a distinct ImageNet class. \textbf{Right:} FID comparison against recent baselines. All numbers use the same evaluation protocol.}
  \label{fig:curated_samples}
\end{figure}

\section{Concluding remarks}
\label{sec:discussion}

We have introduced \emph{Beckmann Transport Models}, a framework for generative modeling built on currents $j=\nu b$ satisfying the divergence equation $\divg j = \mu_0 - \mu_1$. We proved the transport property separately for the finite FM current, for finite-mass weighted Beckmann minimizers with uniform terminal orientation, and for the canonical Coulomb current. The framework includes a Coulomb construction closely related to Poisson Flow and a flow-matching-consistent correction of Equilibrium Matching. The resulting one-step map sends each initial point to the endpoint of its autonomous trajectory and satisfies a conservation law along characteristics, which motivates the direct learning objective evaluated in this paper. For the straight FM and Coulomb constructions, our proofs require the target $\mu_1$ to be supported on a manifold of codimension at least two; in codimension one, counterexamples arise because the current may pass through the target. The general flux-balance result in Appendix~\ref{app:proof} applies in any positive codimension when the terminal direction is imposed explicitly. For targets satisfying the remaining assumptions, zero-padding with at least two coordinates supplies the codimension required by the FM and Coulomb results.

The exact transport statements are population-level results. As in standard diffusion and flow-matching theory, implementations based on finite data and neural networks approximate the population field or map; the ImageNet experiments evaluate the proposed constructions in this practical regime rather than asserting finite-sample exactness.

\paragraph{Open directions.}
One could optimize jointly over admissible pairs $(\nu,b)$ subject to $\divg(\nu b)=\mu_0-\mu_1$, while enforcing~\eqref{eq:no_flux_infinity} and~\eqref{eq:terminal_current}. This could be done in the Beckmann sense, to minimize the static functional $\int_{\mathbb{R}^d}|b|^2\nu\,dx$, or under other criteria targeting sample quality, computational cost, or training stability. For a fixed admissible current $j$, changing $\nu$ changes only the clock $b=j/\nu$ and provides a particularly transparent design direction.

The framework may also be relevant to text generation via embedding on the probability simplex: text tokens correspond to vertices of the simplex (atomic distributions), so the singular-target assumption is satisfied naturally. The conservation equation then provides a stationary condition for learning a map directly to a discrete output, an avenue worth exploring as discrete generative modeling continues to develop.

Finally, the empirical gains from iterated inference in Section~\ref{sec:direct:l} suggest a trade-off between training-time and inference-time compute that has not yet been characterized theoretically. Understanding when parameter-space training and repeated composition approximate the frozen flow-map evolution is an open problem with practical relevance.

\begin{ack}
We thank Shiyi Wang for discussion on presentation. PP is supported by the EPSRC CDT in Modern Statistics and Statistical Machine Learning [EP/S023151/1], a Google PhD Fellowship, and an NSERC Postgraduate Scholarship (PGS D). MSA is supported by a Junior Fellowship at the Harvard Society of Fellows as well as the National Science Foundation under Cooperative Agreement PHY-2019786 (The NSF AI Institute for Artificial Intelligence and Fundamental Interactions). This work has been made possible in part by a gift from the Chan Zuckerberg Initiative Foundation to establish the Kempner Institute for the Study of Natural and Artificial Intelligence.
\end{ack}

\bibliographystyle{plainnat}
\bibliography{refs}

\clearpage
\appendix
\addtocontents{toc}{\protect\setcounter{tocdepth}{2}}
\renewcommand{\contentsname}{Appendix contents}
\pdfbookmark[0]{Appendix contents}{appendix-contents}
\tableofcontents
\clearpage

\section{Proofs for autonomous transport and the endpoint map}
\label{app:proof}

\paragraph{Precise form of Assumption~\ref{ass:source_target}.}
For the proofs below, \emph{rapidly decaying} means that $\rho_0$ is a
Schwartz density. Thus $\rho_0\in C^\infty(\R^d)$ is strictly positive
and, for every multi-index $\beta$ and integer $N\ge0$,
\[
\sup_{x\in\R^d}(1+|x|)^N
|\partial^\beta\rho_0(x)|<\infty.
\]
In particular, $\rho_0$ has moments of every order. It is enough for
$M_1$ to be a compact embedded $C^3$ submanifold without boundary and
for its positive density $\rho_1$ to belong to $C^1(M_1)$. Precisely,
\[
\mu_1=\rho_1\,\mathcal H^k\!\restriction_{M_1},
\qquad q:=d-k\ge2.
\]
Thus the natural volume measure used in the main text is
$\mathrm{vol}_{M_1}=\mathcal H^k\!\restriction_{M_1}$.
Here $\mathcal H^k$ is the $k$-dimensional Hausdorff measure, normalized to agree with Lebesgue measure on $\R^k$: for $A\subset\R^d$,
\[
\mathcal H^k(A)
:=\lim_{\delta\downarrow0}\inf\left\{
\sum_i\alpha_k\left(\frac{\operatorname{diam}U_i}{2}\right)^k:
A\subset\bigcup_iU_i,\quad \operatorname{diam}U_i<\delta
\right\},
\]
where $\alpha_k$ is the volume of the unit ball in $\R^k$. For $k=0$, $\mathcal H^0$ is counting measure. The restriction notation means
$\bigl(\mathcal H^k\!\restriction_{M_1}\bigr)(A)=\mathcal H^k(A\cap M_1)$.

\paragraph{Weighted-current convention in Theorem~\ref{thm:main}.}
For a positive finite weight $\nu$, write
\[
H_\nu:=L^2(\nu^{-1}dx;\R^d),
\qquad
\lVert j\rVert_{H_\nu}^2
:=\int_{\R^d}\frac{|j|^2}{\nu}\,dx.
\]
The divergence of $j\in H_\nu$ is understood distributionally. Indeed,
for every $\zeta\in C_c^\infty(\R^d)$,
\[
\left|\int_{\R^d}j\cdot\nabla\zeta\,dx\right|
\le
\lVert j\rVert_{H_\nu}
\left(\int_{\R^d}\nu|\nabla\zeta|^2\,dx\right)^{1/2}.
\]
It follows that the divergence constraint in
\eqref{eq:beckmann_class} is closed in $H_\nu$. Thus
$\mathcal J_\nu$ is a closed convex subset of a Hilbert space. If it is
nonempty, it contains a unique element of smallest $H_\nu$ norm, which
is the minimizer in \eqref{eq:beckmann_current}. Therefore
Theorem~\ref{thm:main} needs only the stated nonemptiness assumption.

The following theorem is the common transport principle behind the FM,
weighted Beckmann, and Coulomb constructions. It separates the
construction-specific estimates from the dynamical conclusion: the
divergence equation identifies the source and sink, the condition at
infinity prevents loss of mass, and the condition near $M_1$ makes the
target terminal for the autonomous flow. Its proof uses only the
classical change-of-variables and divergence formulas for a $C^1$
flow. Unlike the construction-specific results, this theorem requires
only that the target have positive codimension.

\begin{theorem}[Autonomous transport from flux balance]
\label{thm:flow_balance}
Assume the regularity and measure conditions of
Assumption~\ref{ass:source_target}, except that here it is enough that
$M_1$ have positive codimension, $k\le d-1$. Define
$r(x):=\operatorname{dist}(x,M_1)$, and let
$\nu,b\in C^1(\R^d\setminus M_1)$ with $\nu>0$, set
$j:=\nu b$, and assume that $j\in L^1_{\mathrm{loc}}(\R^d;\R^d)$.
Suppose that, for every $\zeta\in C_c^\infty(\R^d)$,
\begin{equation}
\label{eq:divcond_weak}
\int_{\R^d}j\cdot\nabla\zeta\,dx
=-\int_{\R^d}\zeta\rho_0\,dx
+\int_{M_1}\zeta\rho_1\,d\mathrm{vol}_{M_1},
\end{equation}
and that
\begin{equation}
\label{eq:no_flux_infinity}
\lim_{R\to\infty}\frac1R
\int_{\{R<|x|<2R\}}|j(x)|\,dx=0.
\end{equation}
Assume also that there are $\kappa>0$ and an $r_0>0$ smaller than the
tubular-neighborhood radius of $M_1$ such that
\begin{equation}
\label{eq:terminal_current}
j(x)\cdot\nabla r(x)\le-\kappa|j(x)|<0
\qquad (0<r(x)<r_0).
\end{equation}
Then, for $\mu_0$-almost every initial point, the forward solution of
$\dot X_t=b(X_t)$ has a unique endpoint $T(x)\in M_1$. Moreover,
$T_\sharp\mu_0=\mu_1$. If $\Gamma$ is the arclength parametrization
of the corresponding flow line, its autonomous travel time is
\begin{equation}
\label{eq:hitting_integral_appendix}
\tau(x)=\int_\Gamma\frac{d\ell}{|b|},
\end{equation}
with the value $+\infty$ allowed.
\end{theorem}

When $k=d-1$, the punctured tubular neighborhood has two sides, and
\eqref{eq:terminal_current} must hold on both. Thus the theorem allows
codimension one only when the current points toward $M_1$ from either
side rather than passing through it.

\begin{proof}
We begin with a flow-tube identity. Let $S$ be a smooth
$(d-1)$-dimensional surface crossed by $b$, with unit normal $n$ chosen
so that $b\cdot n>0$. Set $F(t,y):=X_{-t}(y)$ and retain only the backward
segments for which $y$ is the first forward intersection with $S$.
ODE uniqueness makes $F$ one-to-one on the resulting tube. The usual
flow-Jacobian formula gives the volume element
\begin{equation}
\label{eq:flow_tube_jacobian}
dx=(b(y)\cdot n(y))
\exp\!\left(-\int_0^t\divg b(F(s,y))\,ds\right)
\,dt\,d\mathcal H^{d-1}(y),
\qquad x=F(t,y).
\end{equation}
Write the exponential factor as $\Theta(t,y)$. Since
$\divg(\nu b)=\rho_0$ along these trajectories,
\begin{equation}
\label{eq:flow_tube_weight}
-\frac{d}{dt}\bigl[\nu(F(t,y))\Theta(t,y)\bigr]
=\rho_0(F(t,y))\Theta(t,y).
\end{equation}
Integrating \eqref{eq:flow_tube_weight} along each backward segment and
using \eqref{eq:flow_tube_jacobian} shows that the source mass in a
finite tube is its outgoing flux through $S$ minus the nonnegative
incoming flux through the back of the tube. For example, if all
segments are truncated at time $L$, then
\[
\int_{F((0,L)\times S)}\rho_0(x)\,dx
=\int_S(b\cdot n)(y)
\bigl[\nu(y)-\nu(F(L,y))\Theta(L,y)\bigr]
\,d\mathcal H^{d-1}(y).
\]
The subtracted term is nonnegative, so the source mass is at most
\begin{equation}
\label{eq:flow_tube_upper_bound}
\int_S j\cdot n\,d\mathcal H^{d-1}.
\end{equation}
The same calculation can be weighted by a bounded function of the exit
point $y$. A smooth exit surface is handled by a countable collection
of such coordinate pieces; their transverse parts are made disjoint
and the identities are summed. Tangencies have zero normal flux and
are obtained by approximation. We shall use this identity for spheres
and for the level sets of $r$.

We now rule out the two ways in which a positive set of trajectories
could fail to approach $M_1$.

For escape to infinity, the coarea formula and
\eqref{eq:no_flux_infinity} give radii $R_n\uparrow\infty$ such that
\begin{equation}
\label{eq:sphere_flux_vanishes}
\int_{\partial B_{R_n}}|j\cdot n|\,d\mathcal H^{d-1}\longrightarrow0.
\end{equation}
Let $E_n$ be the set of points in $B_{R_n}$ whose forward trajectories
leave $B_{R_n}$ before reaching the terminal tube. Applying
\eqref{eq:flow_tube_upper_bound} to their first-exit points gives
\begin{equation}
\label{eq:escape_flux_bound}
\mu_0(E_n)
\le
\int_{\partial B_{R_n}}|j\cdot n|\,d\mathcal H^{d-1}.
\end{equation}
Once a trajectory enters the terminal tube it cannot leave it, by
\eqref{eq:terminal_current}. Hence every unbounded trajectory either
starts outside $B_{R_n}$ or belongs to $E_n$. Since
$\mu_0(\R^d\setminus B_{R_n})\to0$, equations
\eqref{eq:sphere_flux_vanishes}--\eqref{eq:escape_flux_bound} show that
the set of unbounded forward trajectories is $\mu_0$-null.

Next let $K$ be a compact subset of $\R^d\setminus M_1$ whose distance
from $M_1$ is positive, and let $A_K$ consist of the points whose
forward trajectories remain in $K$. Wherever the flow is defined, its
Jacobian satisfies
\[
\nu(X_t(x))|\det DX_t(x)|
=\nu(x)\exp\!\left(
\int_0^t\frac{\rho_0(X_s(x))}{\nu(X_s(x))}\,ds\right).
\]
The last ratio has a positive minimum $m_K$ on $K$. Hence
\[
\int_{X_t(A_K)}\nu(y)\,dy
\ge e^{m_Kt}\int_{A_K}\nu(x)\,dx.
\]
The left-hand side is at most $\int_K\nu<\infty$, so $A_K$ has zero
$\nu$-measure. Since $\nu>0$, the set $A_K$ has zero Lebesgue measure
and therefore also zero $\mu_0$-measure.

For each positive integer $m$, set
\[
K_m:=\{x:|x|\le m,\ r(x)\ge r_0\}.
\]
A bounded trajectory that never enters $\{r<r_0\}$ remains in one of
the compact sets $K_m$. Each corresponding set $A_{K_m}$ is null by
the preceding argument, and the set of unbounded trajectories was
already shown to be null. Hence almost every trajectory enters
$\{r<r_0\}$. Because $\nu>0$,
\eqref{eq:terminal_current} is equivalent to
\[
b\cdot\nabla r\le-\kappa|b|<0.
\]
Thus, when the part of a trajectory inside the tube is parametrized by
arclength $\ell$, it satisfies $dr/d\ell\le-\kappa$. Its remaining
length is at most $r_0/\kappa$, so the trajectory converges to a single
point. That point must lie on $M_1$: if its distance from $M_1$ were
positive, continuity and the strict inequality $|b|>0$ would force
$r$ to keep decreasing at a uniform positive rate. The inequality also
prevents the trajectory from leaving the tube. This proves existence
and uniqueness of the endpoint outside a $\mu_0$-null set.
Measurability follows by expressing $T$ as the limit of the measurable
first-hitting maps of the levels $\{r=\varepsilon_m\}$ for any sequence
$\varepsilon_m\downarrow0$.

It remains to identify the endpoint law. Fix
$\varepsilon\in(0,r_0)$, let
$S_\varepsilon:=\{r=\varepsilon\}$ and let $H_\varepsilon(x)$ be the
first point at which the trajectory from $x$, with $r(x)>\varepsilon$,
hits $S_\varepsilon$. If $g$ is continuous on $S_\varepsilon$, flux
balance along the flow tubes ending at $S_\varepsilon$ gives
\begin{equation}
\label{eq:first_hit_flux}
\int_{\{r>\varepsilon\}}
g(H_\varepsilon(x))\rho_0(x)\,dx
=
\int_{S_\varepsilon}
g(y)\bigl[-j(y)\cdot\nabla r(y)\bigr]
\,d\mathcal H^{d-1}(y).
\end{equation}
Indeed, apply the weighted flow-tube identity to the backward tubes
starting on $S_\varepsilon$. Along any backward trajectory that remains
in a compact set away from $M_1$, equation
\eqref{eq:flow_tube_weight} and the positive lower bound of
$\rho_0/\nu$ imply $\nu(F(t,y))\Theta(t,y)\to0$. For the unbounded
backward trajectories, truncate the tubes at $\partial B_{R_n}$; the
total incoming term is bounded by the flux in
\eqref{eq:sphere_flux_vanishes} and also tends to zero. Passing to the
limit leaves only the flux through $S_\varepsilon$, which proves
\eqref{eq:first_hit_flux}. The exceptional non-hitting set found above
has zero Lebesgue measure and contributes nothing.

Let $p$ denote nearest-point projection onto $M_1$ in the tubular
neighborhood and take $g=f\circ p$, where $f\in C^1(M_1)$. Applying
\eqref{eq:divcond_weak} to smooth approximations of the indicator of
$\{r<\varepsilon\}$ gives, for almost every $\varepsilon$,
\begin{align}
&\int_{S_\varepsilon}f(p(y))
\bigl[-j(y)\cdot\nabla r(y)\bigr]
\,d\mathcal H^{d-1}(y) \notag\\
&\qquad=
\int_{M_1}f\rho_1\,d\mathrm{vol}_{M_1}
-\int_{\{r<\varepsilon\}}f(p(x))\rho_0(x)\,dx
-\int_{\{r<\varepsilon\}}j(x)\cdot\nabla(f\circ p)(x)\,dx.
\label{eq:tube_flux_limit}
\end{align}
The last two terms tend to zero: the first because $M_1$ has zero
Lebesgue measure, and the second because $j$ is locally integrable and
$\nabla(f\circ p)$ is bounded. On the other hand,
$p(H_\varepsilon(x))\to T(x)$ for almost every $x$, and
$\mu_0(\{r\le\varepsilon\})\to0$. Combining
\eqref{eq:first_hit_flux} and \eqref{eq:tube_flux_limit}, then using
dominated convergence, yields
\[
\int_{\R^d}f(T(x))\,d\mu_0(x)
=\int_{M_1}f\,d\mu_1.
\]
Density extends this identity from $C^1(M_1)$ to continuous functions,
which is exactly $T_\sharp\mu_0=\mu_1$. Finally,
$dt=d\ell/|b|$ along an arclength-parametrized trajectory, proving
\eqref{eq:hitting_integral_appendix}.
\end{proof}

\begin{proof}[Proof of Proposition~\ref{prop:fmtransport}]
\emph{Step 1: The FM minimizer and its flux.}
For $s<1$, let $\rho_s$ be the density of $I_s$ and let
\[
b_s(x):=\E[\dot I_s\mid I_s=x]
\]
be the usual time-dependent FM drift. Define the occupation density and
the time-averaged flux by
\begin{equation}
\label{eq:nu_j_definition}
\nu(x):=\int_0^1\rho_s(x)\,ds,
\qquad
j(x):=\int_0^1\rho_s(x)b_s(x)\,ds.
\end{equation}
Since $\alpha_s>0$ for $s<1$ and $\rho_0$ is strictly positive,
$\rho_s$ and hence $\nu$ are strictly positive.

For any time-independent field $\widehat b$, conditioning on $I_s$ and
also averaging over the uniform variable $s$ gives
\[
\E\bigl[|\widehat b(I_s)-\dot I_s|^2\bigr]
=
\int_{\R^d}\nu(x)
\left|\widehat b(x)-\frac{j(x)}{\nu(x)}\right|^2dx+C,
\]
where $C$ does not depend on $\widehat b$. Therefore the minimizer in
\eqref{eq:FMloss} is
\begin{equation}
\label{eq:b_equals_j_over_nu}
b(x)=\frac{j(x)}{\nu(x)}.
\end{equation}

The divergence equation follows directly from the endpoints of the
interpolant. For every $\varphi\in C_c^\infty(\R^d)$,
\begin{align}
\int_{\R^d}j(x)\cdot\nabla\varphi(x)\,dx
&=\E\int_0^1\nabla\varphi(I_s)\cdot\dot I_s\,ds \notag\\
&=\E[\varphi(x_1)-\varphi(x_0)].
\label{eq:current_boundary}
\end{align}
Since
$\langle\divg j,\varphi\rangle:=-\int_{\R^d} j\cdot\nabla\varphi\,dx$,
this is exactly
\[
\divg j=\mu_0-\mu_1.
\]

Changing variables from $s$ to $a=\alpha_s$ also gives, for every
compactly supported smooth vector field $\psi$,
\begin{equation}
\label{eq:clock_free_current}
\int_{\R^d}\psi(x)\cdot j(x)\,dx
=
\E\int_0^1
\psi\bigl(a x_0+(1-a)x_1\bigr)\cdot(x_1-x_0)\,da.
\end{equation}
Thus $j$ depends on the straight paths, but not on the clock used to
traverse them. Conditional Jensen's inequality also gives
\[
\int_{\R^d}\frac{|j(x)|^2}{\nu(x)}\,dx
=\int_{\R^d}\nu(x)|b(x)|^2\,dx
\le \E\int_0^1|\dot I_s|^2\,ds<\infty.
\]
The last quantity is finite because
$\dot I_s=\dot\alpha_s(x_0-x_1)$, $\dot\alpha$ is bounded, and both
endpoint laws have finite second moments.
Together with the divergence equation, this proves directly that the
FM current belongs to the feasible class $\mathcal J_\nu$, although it
need not be its minimizer. The corresponding first-moment estimate is
\[
\int_{\R^d}|j(x)|\,dx
\le
\int_0^1\int_{\R^d}\rho_s(x)|b_s(x)|\,dx\,ds
\le
\E\int_0^1|\dot I_s|\,ds
=
\E|x_1-x_0|<\infty.
\]

Conditioning on $x_1=y$ and integrating out $x_0$ gives, for
$x\notin M_1$,
\begin{align}
\label{eq:j_straight_density}
j(x)
&=\int_0^1\int_{M_1}
\frac{y-x}{a}\,a^{-d}
\rho_0\!\left(\frac{x-(1-a)y}{a}\right)
\rho_1(y)\,d\mathcal H^k(y)\,da,\\
\label{eq:nu_straight_density}
\nu(x)
&=\int_0^1\int_{M_1}
\frac{1}{-\dot\alpha_{\alpha^{-1}(a)}}\,a^{-d}
\rho_0\!\left(\frac{x-(1-a)y}{a}\right)
\rho_1(y)\,d\mathcal H^k(y)\,da.
\end{align}
The assumptions on $\rho_0$, $\rho_1$, and $M_1$ allow
differentiation under these integrals on compact subsets of
$\R^d\setminus M_1$. Consequently $j$, $\nu$, and $b=j/\nu$ are $C^1$
there.

\emph{Step 2: Terminal geometry near $M_1$.}
Let $p$ be the nearest-point projection of $x$ onto $M_1$ in a tubular neighborhood and write
\[
x=p+r\xi,
\qquad
r=\operatorname{dist}(x,M_1),
\qquad
\xi\in N_pM_1,\qquad |\xi|=1.
\]
Uniformly for $p\in M_1$ and unit normal vectors $\xi\in N_pM_1$,
\begin{equation}
\label{eq:j_terminal_asymptotic}
j(p+r\xi)
=\rho_1(p)r^{1-q}\bigl(F(p,\xi)+o(1)\bigr),
\end{equation}
where
\begin{equation}
\label{eq:F_terminal}
F(p,\xi)
:=\int_0^\infty u^{q-2}
\int_{T_pM_1}(z-u\xi)\rho_0(p+u\xi-z)
\,d\mathcal H^k(z)\,du.
\end{equation}
Indeed, in a local parametrization of $M_1$ set
\[
y=p+az+O(a^2|z|^2),
\qquad z\in T_pM_1,
\]
and then put $u=r/a$. The surface element is
\[
d\mathcal H^k(y)=a^k(1+O(a|z|))\,d\mathcal H^k(z),
\]
while
\[
\frac{y-x}{a}=z-u\xi+o(1),
\qquad
\frac{x-(1-a)y}{a}=p+u\xi-z+o(1).
\]
Since $a^{-d}a^k=a^{-q}$ and $da=-r u^{-2}du$, these terms combine as
$r^{1-q}u^{q-2}$. On bounded sets of $(u,z)$ the displayed expansions
converge uniformly. The rapid decay of $\rho_0$ controls the tails in
$u$ and $z$; it also makes the contribution from outside the coordinate
patch, or from $a$ bounded away from zero, smaller than $r^{1-q}$.
Dominated convergence therefore gives
\eqref{eq:j_terminal_asymptotic}. This is where $q\ge2$ is used: it
makes $u^{q-2}$ integrable at $u=0$ and makes the local term dominate
the bounded remainder.

The normal component of \eqref{eq:F_terminal} is
\begin{equation}
\label{eq:F_inward}
F(p,\xi)\cdot\xi
=-\int_0^\infty u^{q-1}
\int_{T_pM_1}\rho_0(p+u\xi-z)
\,d\mathcal H^k(z)\,du<0.
\end{equation}
The set of pairs $(p,\xi)$ with $p\in M_1$ and $\xi$ a unit normal is
compact. Together with the strict positivity of $\rho_0$, this gives
constants $c_0,C_0,r_0>0$ such that
\begin{equation}
\label{eq:inward_cone}
j(p+r\xi)\cdot\xi\le -c_0r^{1-q},
\qquad
|j(p+r\xi)|\le C_0r^{1-q}
\end{equation}
whenever $0<r<r_0$. Because $b=j/\nu$ with $\nu>0$, the fields $b$ and
$j$ have the same flow lines and direction. If a forward flow line
$\Gamma$ in the tube is parametrized by arclength, then
\[
\frac{d}{d\ell}\operatorname{dist}(\Gamma(\ell),M_1)
\le-\frac{c_0}{C_0}.
\]
Consequently, once a flow line enters this neighborhood, its remaining
length is finite and it converges to one point of $M_1$. The distance
keeps decreasing, so the flow line cannot leave the neighborhood.
For a codimension-one target, the local term need not dominate the
remainder, and this inward estimate does not follow automatically.

\emph{Step 3: Global confinement and endpoint distribution.}
The straight-path formula gives a particularly simple global geometry.
For $x\notin M_1$, define
\[
w_x(a,y):=a^{-d-1}
\rho_0\!\left(\frac{x-(1-a)y}{a}\right)\rho_1(y),
\qquad
\lambda(x):=\int_0^1\int_{M_1}w_x(a,y)
\,d\mathcal H^k(y)\,da,
\]
and
\[
\bar y(x):=\frac1{\lambda(x)}
\int_0^1\int_{M_1}y\,w_x(a,y)
\,d\mathcal H^k(y)\,da.
\]
The integrals are finite, $\lambda(x)>0$, and
\eqref{eq:j_straight_density} becomes
\begin{equation}
\label{eq:fm_convex_hull_field}
j(x)=\lambda(x)\bigl(\bar y(x)-x\bigr),
\qquad
\bar y(x)\in C:=\operatorname{conv}(M_1).
\end{equation}
Let $p_C(x)$ be the nearest point of the compact convex set $C$ and
write $d_C(x)=\operatorname{dist}(x,C)$. For $x\notin C$, the
projection inequality
$(\bar y-p_C(x))\cdot(x-p_C(x))\le0$ gives
\begin{equation}
\label{eq:fm_convex_confinement}
j(x)\cdot\nabla d_C(x)
\le-\lambda(x)d_C(x)<0.
\end{equation}
The same inequality holds with $b=j/\nu$ in place of $j$, up to the
positive factor $1/\nu$. Hence, for every $R>0$, the neighborhood
$\{d_C\le R\}$ is forward invariant: autonomous FM trajectories
cannot escape to infinity.

For the endpoint distribution, the finite-mass estimate in Step~1
implies
\[
\frac1R\int_{\{R<|x|<2R\}}|j(x)|\,dx
\le \frac{\lVert j\rVert_{L^1}}R\longrightarrow0.
\]
Together with \eqref{eq:current_boundary}, the regularity established
in Step~1, and the inward estimate \eqref{eq:inward_cone}, this verifies
the hypotheses of the flow-balance
Theorem~\ref{thm:flow_balance}. It follows that the endpoint exists for
$\mu_0$-almost every initial point and that
$T_\sharp\mu_0=\mu_1$.

\emph{Step 4: Clock dependence and hitting time.}
Equation~\eqref{eq:clock_free_current} shows that $j$ depends only on
the straight-line geometry. The clock changes $\nu$ and therefore the
speed of $b=j/\nu$, but not its flow lines or the map $T$.

Assumption~\ref{ass:fm} implies, as $a\downarrow0$,
\[
-\dot\alpha_{\alpha^{-1}(a)}
=c_\alpha a^\sigma(1+o(1)),
\qquad
\sigma:=\frac{\gamma}{\gamma+1}\in[0,1),
\qquad
c_\alpha:=A^{1/(\gamma+1)}(\gamma+1)^{\gamma/(\gamma+1)}.
\]
Using this relation in \eqref{eq:nu_straight_density} and repeating the
calculation from Step~2 gives
\begin{equation}
\label{eq:nu_terminal_asymptotic}
\nu(p+r\xi)
=\frac{\rho_1(p)}{c_\alpha}r^{1-q-\sigma}
\bigl(G_\sigma(p,\xi)+o(1)\bigr),
\end{equation}
where
\[
G_\sigma(p,\xi)
:=\int_0^\infty u^{q+\sigma-2}
\int_{T_pM_1}\rho_0(p+u\xi-z)
\,d\mathcal H^k(z)\,du>0.
\]
Combining \eqref{eq:j_terminal_asymptotic} and
\eqref{eq:nu_terminal_asymptotic} gives
\begin{equation}
\label{eq:b_terminal_asymptotic}
b(p+r\xi)
=c_\alpha r^\sigma\frac{F(p,\xi)}{G_\sigma(p,\xi)}+o(r^\sigma).
\end{equation}
There are constants $C_1,C_2>0$ such that every trajectory in a
sufficiently small neighborhood of $M_1$ satisfies
\[
-C_2r^\sigma
\le \frac{d}{dt}\operatorname{dist}(X_t,M_1)
\le -C_1r^\sigma.
\]
Since $\sigma<1$,
\[
\int_0^{r_0}\frac{dr}{r^\sigma}<\infty,
\]
so the hitting time is finite. If $\gamma=0$, then
$\dot\alpha_1=-A<0$ and
\eqref{eq:b_terminal_asymptotic}--\eqref{eq:F_inward} give
$|b(x)|\asymp1$ near $M_1$; in general $b$ then has no continuous trace
on $M_1$. If $\gamma>0$, then $\dot\alpha_1=0$, $\dot I_1=0$, and
\eqref{eq:b_terminal_asymptotic} gives
$|b(x)|\asymp r^\sigma=r^{\gamma/(\gamma+1)}\to0$. Extending this field
by zero makes it self-stopping, while $\sigma<1$ still guarantees
finite arrival time.
\end{proof}

\begin{proof}[Proof of Theorem~\ref{thm:main}]
By definition, the minimizing current $j_\nu$ satisfies
$\divg j_\nu=\mu_0-\mu_1$ in the distributional sense of
\eqref{eq:divcond_weak}. By Cauchy--Schwarz, its finite weighted energy
and the finite mass of $\nu$ give
\[
\int_{\R^d}|j_\nu|\,dx
\le
\left(\int_{\R^d}\nu\,dx\right)^{1/2}
\left(\int_{\R^d}\frac{|j_\nu|^2}{\nu}\,dx\right)^{1/2}
<\infty.
\]
Consequently,
\[
\frac1R\int_{\{R<|x|<2R\}}|j_\nu(x)|\,dx
\le \frac{\lVert j_\nu\rVert_{L^1}}R\longrightarrow0,
\]
which is the no-loss condition~\eqref{eq:no_flux_infinity}. Since
condition~\eqref{eq:terminal_beckmann} is precisely
\eqref{eq:terminal_current}, all hypotheses of
Theorem~\ref{thm:flow_balance} therefore hold. The theorem gives the
existence of the endpoint map and
$T_\sharp\mu_0=\mu_1$.

\end{proof}

\begin{proof}[Proof of Theorem~\ref{thm:Tpde}]
Fix $x_0$ outside the exceptional set in
Proposition~\ref{prop:fmtransport}, Theorem~\ref{thm:main}, or
Proposition~\ref{prop:coulomb}. Restarting the autonomous flow from
$X_t(x_0)$ follows the same trajectory with a time shift, so it has the
same endpoint:
\[
T(X_t(x_0))=T(x_0),\qquad 0\le t<\tau(x_0).
\]
At points where $T$ is differentiable, the chain rule gives
$b\cdot\nabla T=0$. Defining $T(p)=p$ for $p\in M_1$ gives the boundary
value.
\end{proof}

\paragraph{Why the boundary value is a terminal condition.}
Theorem~\ref{thm:Tpde} starts with the endpoint map selected by the
flow. It does not claim that the pointwise conditions
$b\cdot\nabla U=0$ off $M_1$ and $U=\mathrm{id}$ on $M_1$ determine a
unique map by themselves. What is needed is that the boundary value be
attained along trajectories approaching $M_1$. More precisely, suppose
that a measurable map $U:\R^d\to M_1$ is constant along
$\mu_0$-almost every trajectory and satisfies
\begin{equation}
\label{eq:terminal_trace}
\lim_{t\uparrow\tau(x)}|U(X_t(x))-X_t(x)|=0.
\end{equation}
Then constancy and the trace condition give
\[
U(x_0)=\lim_{t\uparrow\tau(x_0)}U(X_t)
=\lim_{t\uparrow\tau(x_0)}X_t=T(x_0).
\]
Thus $U=T$ for $\mu_0$-almost every $x_0$. If $U$ is $C^1$, then
$b\cdot\nabla U=0$ implies constancy along trajectories by the chain
rule. The same conclusion holds when $U$ is only absolutely continuous
along the relevant trajectories and the equation holds almost
everywhere on them. Finally, if $U(X_t(x))$ is continuous up to the
endpoint and $U(p)=p$ on $M_1$, then
\eqref{eq:terminal_trace} follows automatically.

\subsection*{Gaussian base and discrete targets}

Let $\mu_0=\mathcal N(0,I_d)$,
$\mu_1=\sum_jp_j\delta_{x_j}$, and
$I_s=\alpha_sx_0+(1-\alpha_s)x_1$. For
$\epsilon:=x-x_j\ne0$, the contribution of atom $j$ to $\nu$ and $j$
is
\begin{align}
\nu^{(j)}(x)
&=\frac{p_j}{(2\pi)^{d/2}}\int_0^1\alpha_s^{-d}
\exp\!\left(-\frac{|\epsilon+\alpha_sx_j|^2}{2\alpha_s^2}\right)ds,
\\
j^{(j)}(x)
&=\frac{p_j}{(2\pi)^{d/2}}\int_0^1\alpha_s^{-d}
\frac{\dot\alpha_s}{\alpha_s}\,\epsilon
\exp\!\left(-\frac{|\epsilon+\alpha_sx_j|^2}{2\alpha_s^2}\right)ds.
\end{align}
The second identity uses $d(1-\alpha_s)/ds=-\dot\alpha_s$. Since
$\dot\alpha_s<0$, each contribution points toward its atom. In
particular, for a single atom every flow line is a straight ray toward
$x_j$, and
\[
b(x)=\lambda(x)(x_j-x),\qquad \lambda(x)>0,
\]
where $\lambda$ depends on $x$ and on the schedule. For
$\alpha_s=(1-s)^a$ with $a\ge1$, Step~4 above gives
$|b(x)|\asymp|x-x_j|^{1-1/a}$ near the atom. Even for $a=1$,
$\lambda$ is generally not identically one.

\section{Learning the first-hitting time}
\label{app:hitting_time_learning}

Motivated by the adaptive stopping in Equilibrium Forcing
\citep{lillemark2026equilibriumforcing}, we consider learning the remaining
first-hitting time directly. Estimating it could allow more precise
sampling of autonomous trajectories by stopping each one when it reaches
the target rather than after a fixed integration horizon. Like the endpoint
map in~\eqref{eq:T_pde}, it obeys an equation along trajectories, but
decreases rather than remaining constant.

\begin{theorem}[Unit decrease of the hitting time]
\label{thm:hitting_time_pde}
Let $b$ be an autonomous drift from Proposition~\ref{prop:fmtransport},
Theorem~\ref{thm:main}, or Proposition~\ref{prop:coulomb}, and suppose its
first-hitting time $\tau(x)$ is finite along the trajectory in question.
Then
\[
\tau(X_t(x))=\tau(x)-t,\qquad 0\le t<\tau(x).
\]
Wherever $\tau$ is differentiable, it follows that
\begin{equation}
\label{eq:hitting_time_pde}
b(x)\cdot\nabla\tau(x)=-1\quad(x\notin M_1),
\qquad \tau(p)=0\quad(p\in M_1).
\end{equation}
\end{theorem}
\begin{proof}
Fix $t<\tau(x)$ and set $y=X_t(x)$. By the semigroup property
\eqref{eq:semigroup}, $X_s(y)=X_{t+s}(x)$ for $0\le s<\tau(x)-t$.
These points lie outside $M_1$ by the definition of the first-hitting
time, whereas
\[
\lim_{s\uparrow\tau(x)-t}X_s(y)
=\lim_{u\uparrow\tau(x)}X_u(x)=T(x)\in M_1.
\]
Hence the trajectory from $y$ first reaches $M_1$ after
$\tau(x)-t$ units of time, i.e. $\tau(X_t(x))=\tau(x)-t$.
At any $x\notin M_1$ where $\tau$ is differentiable, differentiating
this identity from the right at $t=0$ and using $\dot X_0(x)=b(x)$ gives
\[
\nabla\tau(x)\cdot b(x)
=\left.\frac{d}{dt}\tau(X_t(x))\right|_{t=0^+}=-1.
\]
Finally, a trajectory initialized at $p\in M_1$ has zero remaining
hitting time, so $\tau(p)=0$.
\end{proof}

Given the velocity field $b$, the analogue of the map loss~\eqref{eq:Tloss}
uses $b(x)$ directly rather than an FM regression sample $\dot I_s$.

\Needspace{13\baselineskip}
\begin{theorem}[Consistency of the hitting-time stop-gradient update]
\label{thm:hitting_time_objective}
Let $b$ and $\tau$ satisfy Theorem~\ref{thm:hitting_time_pde}, and let $q$
be a sampling distribution on $\R^d\setminus M_1$ for which $\tau$ is
finite and differentiable almost everywhere. Suppose the expectations
below are finite. Then the population stop-gradient update vanishes at
$\tau$ for
\begin{equation}
\label{eq:hitting_time_loss}
\mathcal L_\tau(\tau_\theta)
=\E_{x\sim q}\bigl[\bigl|\tau_\theta(x)-\mathrm{sg}\bigl(
\tau_\theta(x)+1+b(x)\cdot\nabla\tau_\theta(x)
\bigr)\bigr|^2\bigr]
+\lambda\,\E_{x_1\sim\mu_1}\bigl[|\tau_\theta(x_1)|^2\bigr].
\end{equation}
\end{theorem}
\begin{proof}
For an admissible scalar variation $H$, the stopped target is held fixed.
At the true hitting time, the variation of the first term is
\[
-2\E_{x\sim q}\bigl[(1+b(x)\cdot\nabla\tau(x))H(x)\bigr]=0
\]
by Theorem~\ref{thm:hitting_time_pde}; the boundary variation is zero
because $\tau(x_1)=0$ for $\mu_1$-almost every $x_1$.
\end{proof}

The update predicts time remaining directly rather than the noise-level
proxy used by Equilibrium Forcing~\citep{lillemark2026equilibriumforcing};
as with the map loss, it does not guarantee optimization convergence
or select the correct solution without a terminal trace. In practice,
classifier-free guidance may supply a time-independent guided velocity
$b_{\mathrm{CFG}}$ in place of $b$ in~\eqref{eq:hitting_time_loss}.
The same argument applies to its own hitting time if guided trajectories
reach $M_1$ in finite time; guidance alone guarantees neither finite
hitting time nor the target pushforward.

\section{Gradient-constrained flow matching}
\label{app:gradient_fm}

This appendix develops the gradient restriction mentioned in the remark
following Theorem~\ref{thm:main}. It has two equivalent descriptions.
One is ordinary autonomous FM with the parameterization
$b=\nabla\phi$. The other is a Dirichlet problem whose source term is
evaluated only at the two endpoints of each interpolation path. The
latter is especially close to the variational formulation of
\citet{cao2022neis}.

\subsection{The boundary formulation}

The calculation does not require independent endpoints or straight
paths. Let $(x_0,x_1)$ have any coupling with marginals $\mu_0$ and
$\mu_1$, and let $I_t$ be an absolutely continuous interpolant with
$I_0=x_0$ and $I_1=x_1$. Define its occupation weight and current by
\begin{equation}
\label{eq:gradient_occupation_current}
\int_{\R^d}f(x)\nu(x)\,dx
=\E\int_0^1 f(I_t)\,dt,
\qquad
\int_{\R^d}h(x)\cdot j_{\rm FM}(x)\,dx
=\E\int_0^1 h(I_t)\cdot\dot I_t\,dt.
\end{equation}
For every smooth scalar test function $\psi$, the chain rule gives
\begin{equation}
\label{eq:gradient_current_boundary}
\int_{\R^d}j_{\rm FM}\cdot\nabla\psi\,dx
=\E[\psi(x_1)-\psi(x_0)],
\end{equation}
which is the weak identity
$\divg j_{\rm FM}=\mu_0-\mu_1$.

\begin{proposition}[Gradient-constrained FM is a weighted Dirichlet problem]
\label{prop:gradient_fm}
For every $C^1$ potential $\phi$ for which both losses are finite, the
gradient-constrained FM loss and
\begin{equation}
\label{eq:gradient_boundary_loss}
\mathcal E_\nu(\phi)
:=
\frac12\int_{\R^d}\nu|\nabla\phi|^2\,dx
+\int_{\R^d}\phi\,d(\mu_0-\mu_1)
\end{equation}
have exactly the same minimizers. If $\phi$ is a minimizer for which
the first variation is valid, then its current
$j_\nabla:=\nu\nabla\phi$ satisfies
$\divg j_\nabla=\mu_0-\mu_1$ distributionally and is the unique
minimizer of the weighted Beckmann problem~\eqref{eq:beckmann_current}.
\end{proposition}

\begin{proof}
For such a potential, expanding the square and using
\eqref{eq:gradient_occupation_current} gives
\begin{align}
&\frac12\E\int_0^1
\bigl|\nabla\phi(I_t)-\dot I_t\bigr|^2\,dt \nonumber\\
&\qquad=
\frac12\int_{\R^d}\nu|\nabla\phi|^2\,dx
+\int_{\R^d}\phi\,d(\mu_0-\mu_1)
+\frac12\E\int_0^1|\dot I_t|^2\,dt.
\label{eq:gradient_fm_decomposition}
\end{align}
Here the cross term was evaluated with the pathwise identity
\[
\nabla\phi(I_t)\cdot\dot I_t
=\frac{d}{dt}\phi(I_t).
\]
The dot on the right denotes a total derivative along the interpolant;
$\phi$ itself is time-independent.

For every smooth compactly supported $\psi$, differentiating
\eqref{eq:gradient_boundary_loss} at $\phi+\varepsilon\psi$ gives
\[
\int_{\R^d}\nu\nabla\phi\cdot\nabla\psi\,dx
+\int_{\R^d}\psi\,d(\mu_0-\mu_1)=0.
\]
Equivalently,
\begin{equation}
\label{eq:weighted_poisson_gradient}
\divg(\nu\nabla\phi)=\mu_0-\mu_1.
\end{equation}
This implies that $j_\nabla$ is feasible in
\eqref{eq:beckmann_current}.
Let $\widetilde j\in\mathcal J_\nu$ be another finite-energy current
and set $h:=\widetilde j-j_\nabla$. Then $h$ has finite weighted energy
and $\divg h=0$. The same integration-by-parts identity, first for
smooth compactly supported approximations and then by weighted-energy
approximation, gives
\[
\int_{\R^d}\frac{j_\nabla\cdot h}{\nu}\,dx
=\int_{\R^d}\nabla\phi\cdot h\,dx=0,
\]
and hence
\[
\int_{\R^d}\frac{|\widetilde j|^2}{\nu}\,dx
=
\int_{\R^d}\frac{|j_\nabla|^2}{\nu}\,dx
+\int_{\R^d}\frac{|h|^2}{\nu}\,dx.
\]
Thus $j_\nabla$ is the unique minimizer of
\eqref{eq:beckmann_current}.
\end{proof}

The result also gives a convenient existence statement in the natural
homogeneous weighted Sobolev space. Indeed,
\eqref{eq:gradient_current_boundary} and Cauchy--Schwarz show that the
linear functional
\begin{equation}
\label{eq:gradient_linear_bound}
\ell(\psi):=\int\psi\,d(\mu_0-\mu_1)
=-\int j_{\rm FM}\cdot\nabla\psi\,dx
\end{equation}
is bounded by
\[
|\ell(\psi)|
\le
\left(\int_{\R^d}\frac{|j_{\rm FM}|^2}{\nu}\,dx\right)^{1/2}
\left(\int_{\R^d}\nu|\nabla\psi|^2\,dx\right)^{1/2}.
\]
Since the FM current has finite weighted energy, the Riesz theorem
therefore yields a unique weak minimizer modulo an additive constant in
the completion of smooth potentials for the seminorm
$\|\nabla\phi\|_{L^2(\nu dx)}$. Additional regularity is needed to use
its gradient as the $C^1$ vector field required by
Theorem~\ref{thm:main}.

\paragraph{Relation with NEIS.}
For smooth densities on a bounded domain $\Omega$ or a torus,
\citet{cao2022neis} consider
\[
\min_V\int_\Omega
\left\{\frac12\mathcal D|\nabla V|^2
+V(\rho_1-\rho_0)\right\}\,dx,
\qquad
\divg(\mathcal D\nabla V)=\rho_1-\rho_0.
\]
Taking $V=-\phi$ and $\mathcal D=\nu$ turns this exactly into
\eqref{eq:gradient_boundary_loss} and
\eqref{eq:weighted_poisson_gradient}. The purposes and hypotheses are
different: their potential produces a zero-variance NEIS dynamics for
regular densities, whereas here $\nu$ is fixed by the FM interpolant
and $\mu_1$ may be singular. In particular, the endpoint term in
\eqref{eq:gradient_boundary_loss} is estimated directly from paired FM
samples and does not require evaluating a target density.

\paragraph{Sampling the loss.}
With $t\sim\mathrm{Unif}[0,1]$, a minibatch estimator of
\eqref{eq:gradient_boundary_loss} is
\begin{equation}
\label{eq:gradient_boundary_minibatch}
\widehat{\mathcal E}_\nu(\phi_\theta)
=\frac1B\sum_{n=1}^B
\left\{
\frac12|\nabla\phi_\theta(I_{t_n}^{(n)})|^2
+\phi_\theta(x_0^{(n)})-\phi_\theta(x_1^{(n)})
\right\}.
\end{equation}
It has the same population minimizer as regressing
$\nabla\phi_\theta(I_t)$ on $\dot I_t$, but it does not require
evaluating the latter. It still requires spatial derivatives of the
potential; differentiating it with respect to network parameters
therefore involves mixed parameter--space derivatives. Smooth
parameterizations are also natural if one wishes to invoke the
classical ODE results in the main text.

If clocks are sampled from a nonuniform density $q$, importance
weighting the time integral by $1/q(t)$ preserves
\eqref{eq:gradient_boundary_minibatch}. Without that weighting, the
cross term is not purely an endpoint term: for an absolutely continuous
$q$,
\begin{equation}
\label{eq:nonuniform_clock_boundary}
-\int_0^1q(t)\frac{d}{dt}\phi(I_t)\,dt
=q(0)\phi(x_0)-q(1)\phi(x_1)
+\int_0^1q'(t)\phi(I_t)\,dt.
\end{equation}

\subsection{Transport and potential regularity}

The gradient restriction preserves the divergence equation but does
not generally preserve the FM field. More precisely,
\[
\divg(j_{\rm FM}-j_\nabla)=0.
\]
The gradient current is the weighted orthogonal projection of the FM
current onto gradient currents. It therefore has the same source and
sink, but can have different flow lines and a different autonomous
clock.

\paragraph{Why the gradient case is transparent.}
Along every classical trajectory of $\dot X_t=\nabla\phi(X_t)$,
\begin{equation}
\label{eq:gradient_lyapunov}
\frac{d}{dt}\phi(X_t)=|\nabla\phi(X_t)|^2\ge0.
\end{equation}
Thus there are no nonconstant closed trajectories. In the finite-mass
setting relevant here, finite weighted energy also gives
\[
\int_{\R^d}|j_\nabla|\,dx
\le
\left(\int_{\R^d}\nu|\nabla\phi|^2\,dx\right)^{1/2}
\left(\int_{\R^d}\nu\,dx\right)^{1/2}<\infty.
\]
Consequently, $j_\nabla$ satisfies the no-loss condition at infinity in
Theorem~\ref{thm:flow_balance}. More importantly, away from $M_1$ the
weighted Poisson equation reads
$\divg(\nu\nabla\phi)=\rho_0$. Wherever the flow is defined, its
Jacobian therefore obeys
\begin{equation}
\label{eq:gradient_weighted_volume}
\nu(X_t(x))|\det DX_t(x)|
=\nu(x)\exp\!\left(
\int_0^t\frac{\rho_0(X_s(x))}{\nu(X_s(x))}\,ds
\right).
\end{equation}
If a positive-measure family of trajectories remained in a compact set
$K\Subset\R^d\setminus M_1$, the right-hand side would grow at least
exponentially, whereas the left-hand side integrates to at most
$\int_K\nu$. Thus such a family is null. This is the precise version
of the intuition that positive divergence prevents trapping away from
the target; it is stronger than the local observation that a $C^2$
critical point satisfies $\Delta\phi=\rho_0/\nu>0$ and hence cannot be
a local maximum of $\phi$.

What remains to be checked is the behavior at the target. The gradient
ansatz alone does not imply the terminal-orientation condition
\eqref{eq:terminal_beckmann}. In codimension one, choose a unit normal
$n$ and write $j_\nabla^\pm$ for the one-sided normal traces, with the
$+$ side pointed to by $n$. The divergence equation gives only the jump
condition
\[
j_\nabla^+\cdot n-j_\nabla^-\cdot n=-\rho_1.
\]
To make $M_1$ terminal from both sides, one instead needs
$j_\nabla^+\cdot n<0$ and $j_\nabla^-\cdot n>0$. The jump condition does
not imply these two inequalities, so it does not rule out a current
that passes through $M_1$. Once terminal orientation and the stated
regularity hold, Theorem~\ref{thm:flow_balance} applies and yields
$T_\sharp\mu_0=\mu_1$.

\paragraph{Away from the target.}
On every compact $K\Subset\R^d\setminus M_1$, the weighted Poisson
equation is an ordinary uniformly elliptic equation as soon as $\nu$ is
bounded above and below on $K$. Thus, for example,
$\nu\in C^{m,\alpha}(K)$ and $\rho_0\in C^{m-1,\alpha}(K)$ give
$\phi\in C^{m+1,\alpha}(K)$, after fixing an additive constant. In
particular, smooth positive coefficients give a smooth potential away
from the target.

\paragraph{Trace on the singular target.}
For the straight FM occupation weight, Step~4 of the proof of
Proposition~\ref{prop:fmtransport} gives, with
$q=d-k$ and $\sigma=\gamma/(\gamma+1)$,
\[
\nu(p+r\xi)\asymp r^{1-q-\sigma}
\qquad (r\downarrow0).
\]
Since tubular volume is comparable to $r^{q-1}\,dr$ times volume on
$M_1$ and the normal sphere, the local weighted energy is governed by
\begin{equation}
\label{eq:gradient_normal_energy}
\int_0^\varepsilon r^{-\sigma}
|\partial_r\phi|^2\,dr.
\end{equation}
Because $0\le\sigma<1$, this controls a trace on $M_1$. Formally, on
each normal ray, Cauchy--Schwarz gives
\begin{equation}
\label{eq:gradient_trace_estimate}
|\phi(p+r\xi)-\phi|_{M_1}(p)|^2
\le
\frac{r^{1+\sigma}}{1+\sigma}
\int_0^r s^{-\sigma}|\partial_s\phi(p+s\xi)|^2\,ds.
\end{equation}
The corresponding weighted Sobolev trace justifies the term
$\phi(x_1)$ in~\eqref{eq:gradient_boundary_loss}; for a neural
potential it is simply its value evaluated on the target samples.

Finite energy alone does not determine the pointwise terminal profile.
If, in addition, a regular solution has the local flux scaling and
orientation
\begin{equation}
\label{eq:gradient_terminal_profile_assumption}
\nabla\phi(p+r\xi)=r^\sigma
\bigl(B(p,\xi)+o(1)\bigr),
\qquad
B(p,\xi)\cdot\xi<0,
\end{equation}
uniformly in $(p,\xi)$, then integration in the normal direction gives
\begin{equation}
\label{eq:gradient_potential_profile}
\phi(p+r\xi)=\phi|_{M_1}(p)
+r^{1+\sigma}\Phi(p,\xi)+o(r^{1+\sigma}).
\end{equation}
For $\gamma>0$, hence $0<\sigma<1$, this is a
$C^{1,\sigma}$-type extension in the normal variables and
$\nabla\phi\to0$ at the target. This does \emph{not} make the hitting
time infinite: terminal orientation gives
$-\dot r\asymp r^\sigma$, and hence
\[
\int_0^{r_0}\frac{dr}{r^\sigma}<\infty.
\]
The point is that $C^{1,\sigma}$ regularity of $\phi$ makes
$\nabla\phi$ only $\sigma$-H\"older, not Lipschitz. By contrast, if
$\phi$ were $C^{1,1}$ near $M_1$, then
$|\nabla\phi|\lesssim r$ and a trajectory starting away from $M_1$
could approach it only in infinite time. For $\gamma>0$, the profile
above gives $\nabla\phi\to0$; if the trace is differentiable along
$M_1$, its tangential derivative therefore vanishes, so the trace is
constant on each connected component. For $\gamma=0$, one has
$\sigma=0$ and the generic local behavior is conical: $\phi$ is
Lipschitz but need not be differentiable on $M_1$. Terminal orientation
alone does not then rule out a nonzero tangential component or a
nonconstant trace.

\paragraph{Contrast with the Coulomb choice.}
The preceding regular behavior depends essentially on the singular FM
weight. For $\nu\equiv1$, the potential solves
$\Delta\phi=\mu_0-\mu_1$. Near a codimension-$q$ target its singular
part has the familiar normal behavior
\[
\phi\asymp
\begin{cases}
\log r, & q=2,\\
r^{2-q}, & q>2,
\end{cases}
\qquad
|\nabla\phi|\asymp r^{1-q}.
\]
It therefore has neither a finite target trace nor finite Dirichlet
energy near $M_1$ when $q\ge2$. This is why the Coulomb construction is
handled separately in Proposition~\ref{prop:coulomb}, while the
boundary loss~\eqref{eq:gradient_boundary_loss} is naturally matched
to the finite-mass FM occupation weight.
\section[Link with Poisson flows when the weight is one]{Link with Poisson flows when $\nu = 1$}
\label{app:coulomb}

\begin{proof}[Proof of Proposition~\ref{prop:coulomb}]
The distributional identity
\[
\divg\!\left(\frac{x}{\omega_d|x|^d}\right)=\delta_0
\]
shows that~\eqref{eq:coulomb_general} satisfies
\eqref{eq:divcond_weak}. The field is $C^1$ on
$\R^d\setminus M_1$ and locally integrable across $M_1$. Indeed, the
estimate below gives size $r^{1-q}$ near the target, while the volume
element in the normal directions is $r^{q-1}\,dr$.

Because $\mu_0$ and $\mu_1$ have the same total mass, the leading terms
at infinity cancel. Expanding the kernel, using the finite first
moments, and controlling the rapidly decaying tail of $\mu_0$ gives
\begin{equation}
\label{eq:coulomb_far_field}
|b(x)|=O(|x|^{-d})
\qquad (|x|\to\infty).
\end{equation}
It follows that $\int_{\{R<|x|<2R\}}|b|\,dx=O(1)$, which verifies~\eqref{eq:no_flux_infinity}.

It remains to check the direction near the target. Write $x=p+r\xi$,
where $p\in M_1$ and $\xi\in N_pM_1$ is a unit normal. In local tangent
coordinates $y=p+z+O(|z|^2)$, the leading normal component of the
$\mu_1$ term in~\eqref{eq:coulomb_general} is
\[
-\frac{\rho_1(p)r}{\omega_d}
\int_{T_pM_1}\frac{dz}{(r^2+|z|^2)^{d/2}}.
\]
After the substitution $z=rw$, the tangential part of the leading
integral vanishes by symmetry, while the normal part is finite because
$q\ge2$. Uniformly in $p$ and $\xi$, this gives
\begin{equation}
\label{eq:coulomb_terminal_asymptotic}
b(p+r\xi)
=-c_{d,k}\rho_1(p)r^{1-q}\xi+o(r^{1-q}),
\qquad
c_{d,k}:=\frac1{\omega_d}
\int_{\R^k}\frac{dw}{(1+|w|^2)^{d/2}}>0.
\end{equation}
The tangential component, the contribution from the rest of $M_1$,
and the smooth $\mu_0$ field are all smaller than $r^{1-q}$.
Uniformity in $p$ and $\xi$, together with positivity of $\rho_1$,
therefore implies
\eqref{eq:terminal_current}. Together with the divergence identity and
the far-field estimate, Theorem~\ref{thm:flow_balance} now gives
$T_\sharp\mu_0=\mu_1$. Moreover,
\[
-b\cdot\nabla r\asymp r^{1-q},
\qquad
\int_0^{r_0}r^{q-1}\,dr<\infty,
\]
so~\eqref{eq:hitting_integral} gives finite arrival time.

The same asymptotic explains why this result is not covered by the
finite-energy Beckmann theorem with $\nu\equiv1$. In tubular
coordinates,
\[
\int_{\{r<\varepsilon\}}|b|^2\,dx
\asymp
\int_0^\varepsilon r^{2(1-q)}r^{q-1}\,dr
=
\int_0^\varepsilon r^{1-q}\,dr
=\infty
\qquad(q\ge2).
\]
\end{proof}

For the standard Gaussian base, the background term in the canonical
Coulomb field is explicit:
\begin{equation}
\label{eq:bcoulomb}
b(x)
=\underbrace{\frac{x}{\omega_d|x|^d}
\frac{\gamma(d/2,|x|^2/2)}{\Gamma(d/2)}}_{b_0(x)}
-\underbrace{\frac1{\omega_d}\int_{M_1}
\frac{x-y}{|x-y|^d}\,\mu_1(dy)}_{b_1(x)},
\end{equation}
where $\omega_d=2\pi^{d/2}/\Gamma(d/2)$ is the surface area of the
unit sphere in $\R^d$, and $\gamma(a,z)$ is the lower incomplete gamma
function.

In the atomic case $\mu_1=\sum_{j\in[N]}p_j\delta_{x_j}$, the second
term in~\eqref{eq:bcoulomb} is also explicit:
\[
b_1(x)
=\frac1{\omega_d}\int_{M_1}\frac{x-y}{|x-y|^d}\,\mu_1(dy)
=\frac1{\omega_d}\sum_{j\in[N]}p_j
\frac{x-x_j}{|x-x_j|^d}.
\]
At each ODE step this sum can be estimated with a fresh i.i.d.\
mini-batch from $\mu_1$, without training a neural network. This target
term is the Coulomb field estimated in the Poisson Flow construction of
\citet{xu2022pfgm}; the full field in~\eqref{eq:bcoulomb} additionally
contains the base-source term $b_0$.

\section[Proof of the FM action bound]{Proof of the FM action bound~\eqref{eq:fm_action_bound}}
\label{app:cost_bound}

Recall that
\[
j(x)=\int_0^1\rho_s(x)b_s(x)\,ds,
\qquad
\nu(x)=\int_0^1\rho_s(x)\,ds,
\qquad
b(x)=\frac{j(x)}{\nu(x)}.
\]
For each $x$, Jensen's inequality with probability weights
$\rho_s(x)\,ds/\nu(x)$ gives
\[
|b(x)|^2
\le\frac1{\nu(x)}\int_0^1|b_s(x)|^2\rho_s(x)\,ds.
\]
Multiplying by $\nu(x)$ and integrating yields
\[
\int_{\R^d}\nu|b|^2\,dx
\le\int_0^1\int_{\R^d}|b_s|^2\,d\mu_s\,ds.
\]
Moreover, $\int_{\R^d}\nu\,dx=1$, and the divergence equation makes $j$ an
admissible competitor in Beckmann's problem. Hence Cauchy--Schwarz and
the Beckmann characterization of $W_1$ give
\[
W_1^2(\mu_0,\mu_1)
\le\left(\int_{\R^d}|j|\,dx\right)^2
\le\int_{\R^d}\frac{|j|^2}{\nu}\,dx
=\int_{\R^d}\nu|b|^2\,dx.
\]
Combining the last two displays proves
\eqref{eq:fm_action_bound}.

The original time-dependent pair $(\mu_s,b_s)$ is admissible for the
Benamou--Brenier problem, so its action also satisfies
\[
W_2^2(\mu_0,\mu_1)
\le
\int_0^1\int_{\R^d}|b_s|^2\,d\mu_s\,ds.
\]
However, we do not identify $\nu$ with the occupation density of the
autonomous flow: equality of their divergence equations does not by
itself imply equality of the densities. Consequently the
Benamou--Brenier inequality does not give a comparison between
$W_2^2$ and the static quadratic functional
$\int_{\R^d}|j|^2/\nu\,dx$.

\section{Changing the interpolant clock and geometry}
\label{app:clock_geometry}

The clock and the path geometry play different roles. This section
records the distinction used in the main text.

\subsection{Changing only the clock}
\label{sec:invariance}

Consider a fixed family of paths
\[
I_s=\alpha_s x_0+f(\alpha_s)x_1,
\]
where $\alpha\in C^1([0,1])$ decreases strictly from one to zero and
$f\in C^1([0,1])$ satisfies $f(1)=0$, $f(0)=1$. For a compactly
supported smooth vector field $\psi$, the integrated flux satisfies
\[
\int_{\R^d}\psi(x)\cdot j(x)\,dx
=\E\int_0^1\psi(I_s)\cdot\dot I_s\,ds.
\]
Changing variables from $s$ to $a=\alpha_s$ gives
\begin{equation}
\label{eq:j_geometric}
\int_{\R^d}\psi(x)\cdot j(x)\,dx
=-\E\int_0^1
\psi\bigl(a x_0+f(a)x_1\bigr)
\cdot\bigl(x_0+f'(a)x_1\bigr)\,da.
\end{equation}
The right-hand side contains the path $a x_0+f(a)x_1$, but not the
speed at which that path is traversed. Thus changing the clock
$\alpha_s$ without changing the path leaves the current $j$ unchanged.

The occupation density does depend on the clock. If
$\widetilde\rho_a$ is the density of
$a x_0+f(a)x_1$, then
\begin{equation}
\label{eq:nu_kinematic}
\nu(x)
=\int_0^1
\frac{\widetilde\rho_a(x)}
{-\dot\alpha_{\alpha^{-1}(a)}}\,da.
\end{equation}
Consequently a clock change modifies $b=j/\nu$ and hence the speed of
the autonomous flow. Because $\nu>0$, it does not change its flow
lines. Whenever those lines have unique endpoints, their endpoint map
is therefore unchanged. This is the clock invariance used in
Proposition~\ref{prop:fmtransport}.

\subsection{Changing the path geometry}

Changing $f$ is different: it changes the current in
\eqref{eq:j_geometric}, so it may change both the flow lines and their
endpoints. The identity
\[
\divg j=\mu_0-\mu_1
\]
still follows from the endpoints of the interpolant, provided the
current is well defined. That identity alone, however, does not ensure
that the flow reaches $M_1$. One must also check that no positive
amount of current is lost at infinity and that, near $M_1$, the current
points toward the manifold as in
\eqref{eq:terminal_current}. Under the regularity assumptions used
here, Theorem~\ref{thm:flow_balance} turns these checks into the endpoint
transport conclusion.

For the straight paths of Assumption~\ref{ass:fm}, these checks were
carried out in Steps~1--3 of the proof of
Proposition~\ref{prop:fmtransport}. In particular,
\eqref{eq:j_terminal_asymptotic}--\eqref{eq:F_inward} show that the
normal component of $j$ points uniformly toward $M_1$. The clock then
determines only the terminal speed:
\[
|b(x)|\asymp
\operatorname{dist}(x,M_1)^{\gamma/(\gamma+1)}.
\]
Thus $\gamma=0$ gives a nonzero limiting speed, whereas $\gamma>0$
gives a drift that tends to zero. Both cases have finite hitting time.
For another path geometry, the same conclusion is valid only after
checking these two properties directly.

\section{Self-stopping interpolant for BTM image experiments}
\label{app:self-stopping-interpolant}

For the image experiments we use an interpolant designed so that its
population autonomous drift tends to zero at $M_1$. Under the
assumptions of Proposition~\ref{prop:fmtransport}, a trajectory reaches
$M_1$ in finite autonomous time and then remains fixed under the
freezing convention.

For a breakpoint $s_c\in(0,1)$, set
\begin{align}
\label{eq:self-stopping-interpolant}
\alpha_s =
\begin{cases}
\displaystyle 1-\dfrac{2s}{1+s_c},
&0\le s\le s_c,\\[6pt]
\displaystyle \dfrac{(1-s)^2}{1-s_c^2},
&s_c<s\le1.
\end{cases}
\end{align}
The two branches have the same value and derivative at $s_c$, so
$\alpha\in C^1([0,1])$. The schedule is linear before $s_c$ and
quadratic near the endpoint. Hence $\dot I_1=0$ and, under the
assumptions of Proposition~\ref{prop:fmtransport},
$|b(x)|\asymp\operatorname{dist}(x,M_1)^{1/2}$ near $M_1$.
The hitting time remains finite because
$\int_0^\epsilon r^{-1/2}\,dr<\infty$, while the zero extension of $b$
keeps the trajectory fixed after arrival. As $s_c\uparrow1$, the
schedule approaches the linear interpolant; as $s_c\downarrow0$, it
approaches the quadratic schedule.

\section{Comparison with Drifting} \label{app:drifting}
BTM and the Drifting framework of \citet{deng2026drifting} both use a
fixed-point viewpoint, but construct their dynamics differently.
Drifting learns a McKean--Vlasov dynamics intended to move
$\mu_0$ toward a target distribution $\mu_*$ as $t\to\infty$:
\begin{align}
\dot x_t
=V[\mu_t,\mu_*](x_t)
:=\frac{\E[\mathcal K(x_t,X_t)X_t]}{\E[\mathcal K(x_t,X_t)]}
-\frac{\E[\mathcal K(x_t,X_*)X_*]}{\E[\mathcal K(x_t,X_*)]},
\qquad x_0\sim\mu_0,
\end{align}
where $\mathcal K$ is a chosen kernel, $X_t\sim\mu_t$, and
$X_*\sim\mu_*$. During training, the ratios of expectations are
estimated from mini-batches; ratios of empirical averages are generally
biased.

\begin{figure}
    \centering
    \includegraphics[width=1\linewidth]{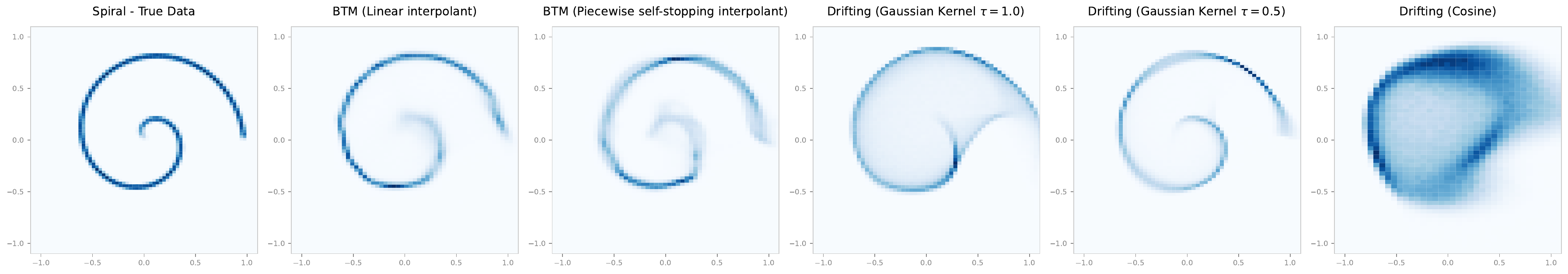}
    \caption{\textbf{Comparison between BTM and Drifting
    \citep{deng2026drifting}.} In this experiment, Drifting is sensitive
    to the kernel choice, whereas BTM trains stably with both
    interpolants shown.}
    \label{fig:BTM-vs-drifting}
\end{figure}

Kernel performance can be sensitive to scale and geometry in high
dimensions, which may require kernel mixtures or temperature tuning.
Figure~\ref{fig:BTM-vs-drifting} compares this behavior with BTM in our
experimental setting. The linear and piecewise self-stopping
interpolants both train without kernel tuning, whereas the Drifting
results vary with the kernel and its hyperparameters.

\section{Additional experiments}
\label{app:experiments}

\begin{figure}[h]
  \centering
  \includegraphics[width=0.45\textwidth]{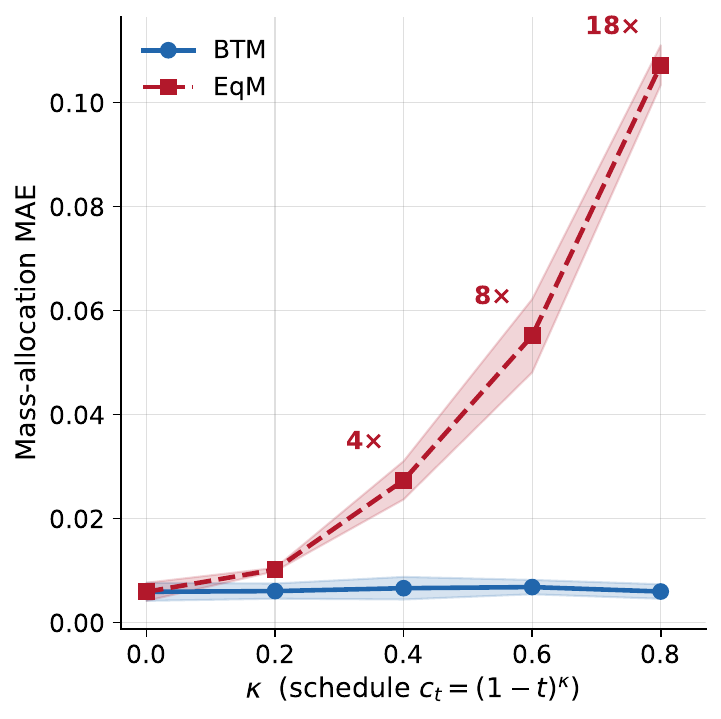}
  \hfill
  \includegraphics[width=0.45\textwidth]{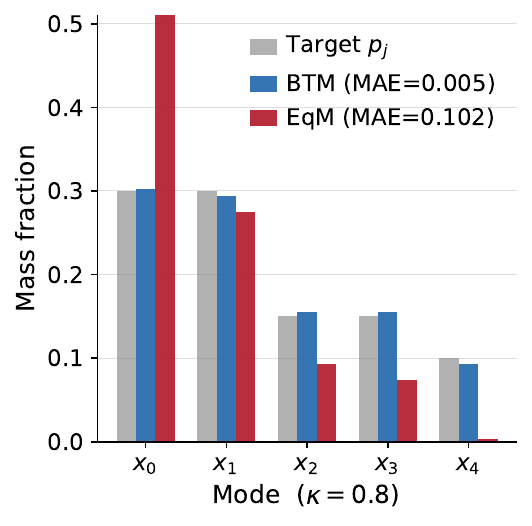}
  \caption{Left: mass-allocation MAE versus schedule exponent $\kappa$
  (mean $\pm$ standard deviation over three seeds). Right: per-atom
  mass fractions at $\kappa=0.8$. The consistent loss (blue) remains
  nearly flat, while the EqM loss (red) degrades by up to a factor of
  $25$.}
  \label{fig:alpha_ablation}
\end{figure}

\paragraph{Bias grows monotonically with $\kappa$ (Figure~\ref{fig:alpha_ablation}).}
Here $c_s=(1-s)^\kappa$ is the schedule multiplying the EqM regression
target. At $\kappa=0$ the two losses are algebraically identical and
both give $\mathrm{MAE}\approx0.007$. As $\kappa$ increases, the EqM
loss degrades monotonically (by up to a factor of $25$ at
$\kappa=0.9$), while the consistent FM loss remains nearly flat.

\begin{figure}[h!]
  \centering
  \includegraphics[width=0.86\textwidth]{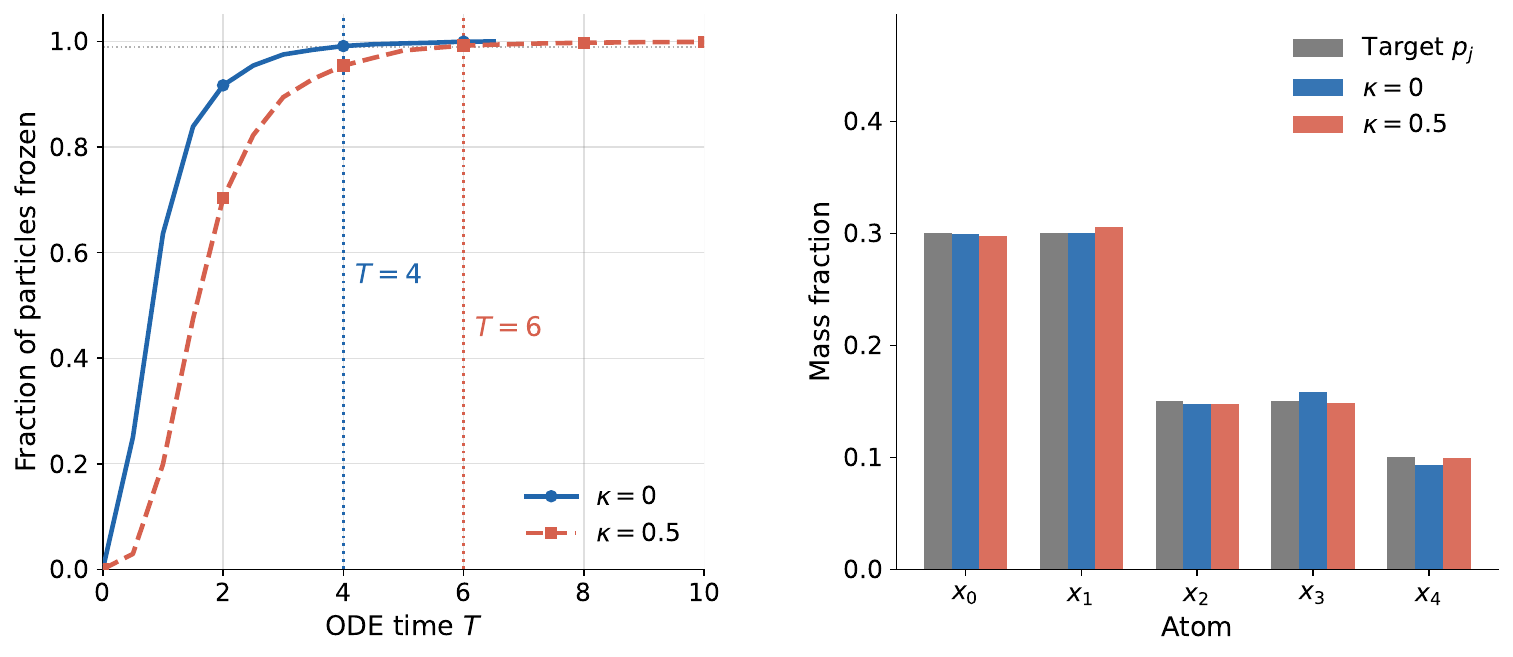}
  \caption{Left: fraction of frozen particles versus ODE time for
  $a=1$ and $a=2$ (equivalently, terminal drift exponents $\sigma=0$
  and $\sigma=1/2$, respectively). Right: both achieve accurate mass allocation
  ($\mathrm{MAE}\approx0.005$); only the convergence speed differs.}
  \label{fig:finite_hitting_time}
\end{figure}

\paragraph{Convergence speed vs.\ correctness (Figure~\ref{fig:finite_hitting_time}).}
For $\alpha_s=(1-s)^a$, both the linear interpolant ($a=1$) and the
quadratic, self-stopping interpolant ($a=2$) have finite autonomous
hitting time. When trained with the consistent loss, both achieve
accurate mass allocation ($\mathrm{MAE}\approx0.005$). They differ
mainly in speed: $99\%$ of particles freeze by autonomous time
$t\approx4$ for $a=1$ and by $t\approx6$ for $a=2$.

\begin{figure}[h!]
  \centering
  \includegraphics[width=0.55\textwidth]{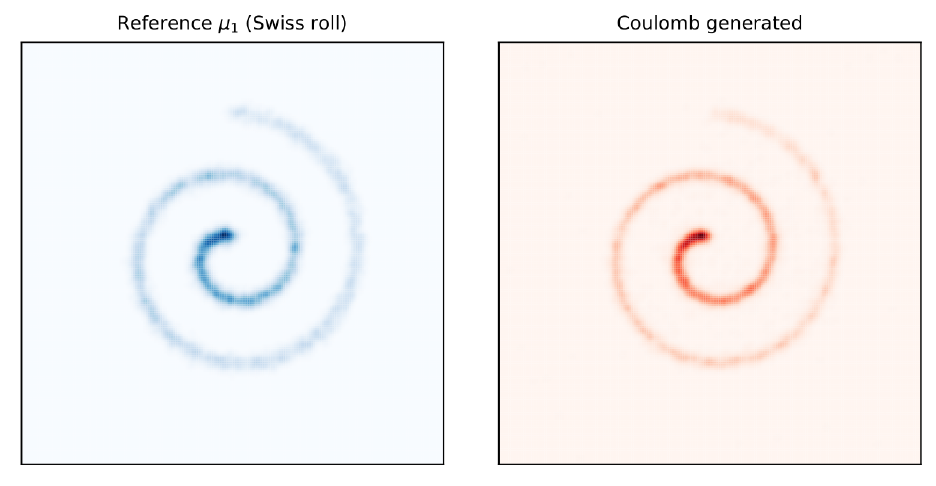}
  \caption{Kernel density estimate of reference $\mu_1$ (left) and Coulomb-transported
  particles (right). We use $\mu_0=\mathcal N(0,I_5)$, $N=5000$
  particles, mini-batches of size $1000$, and integration time $t=30$.
  No training or
  score function is used.}
  \label{fig:coulomb_spiral}
\end{figure}

\paragraph{Training-free Coulomb transport (Figure~\ref{fig:coulomb_spiral}).}
We also test the training-free Coulomb field ($\nu=1$,
Proposition~\ref{prop:coulomb}) on a continuous target. In
Figure~\ref{fig:coulomb_spiral}, mini-batch estimates of the Coulomb
drift transport a Gaussian cloud toward a Swiss-roll target in $d=5$
without a neural network or training loop. Near a target of codimension
$q$, the exact population field has scale $r^{1-q}$, as shown in
\eqref{eq:coulomb_terminal_asymptotic}.

\section{Image experiments for direct map learning}
\label{app:prelim-image}
We report two direct-map experiments: (i) class-conditional MNIST digit
generation and (ii) class-conditional ImageNet $256\times256$
generation in a latent space. MNIST provides a small-scale feasibility
check, while ImageNet is used to study stabilization and model scaling.

\subsection{MNIST digit generation}
\begin{figure}[h!]
  \centering
  \begin{subfigure}[b]{0.48\textwidth}
    \centering
    \includegraphics[width=\linewidth]{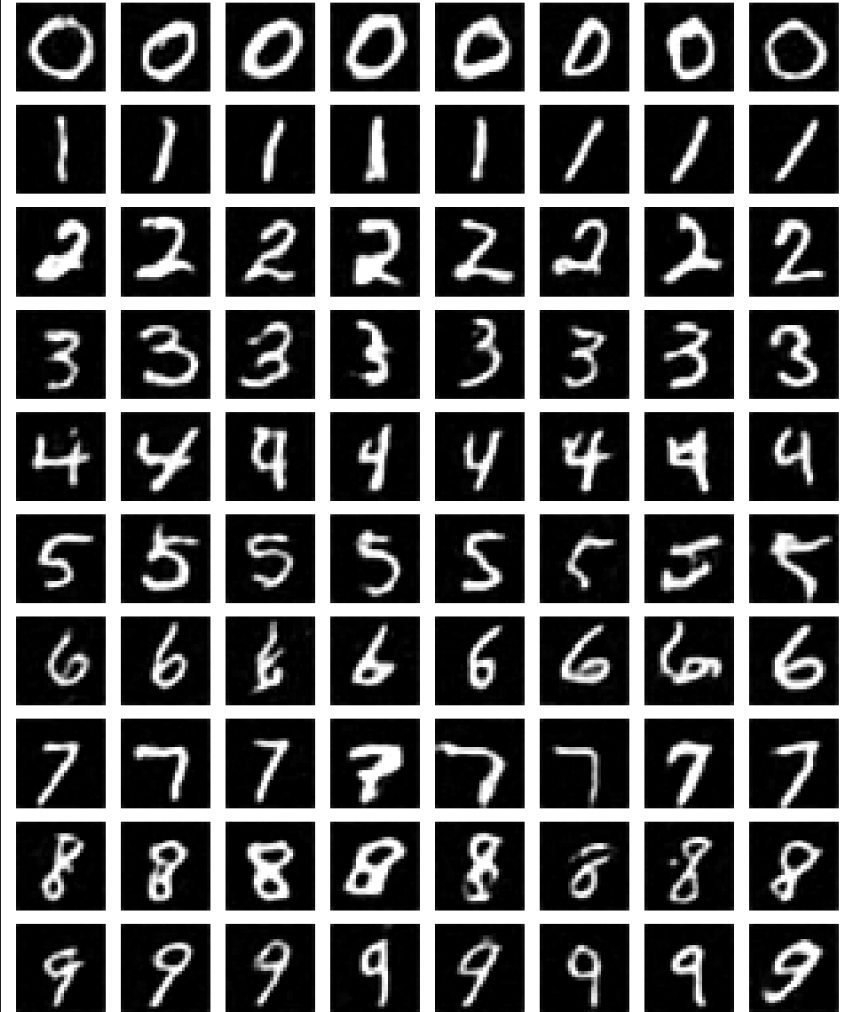}
    \caption{One-step generated results}
    \label{fig:map_mnist_1step}
  \end{subfigure}
  \hfill
  \begin{subfigure}[b]{0.47\textwidth}
    \centering
    \includegraphics[width=\linewidth]{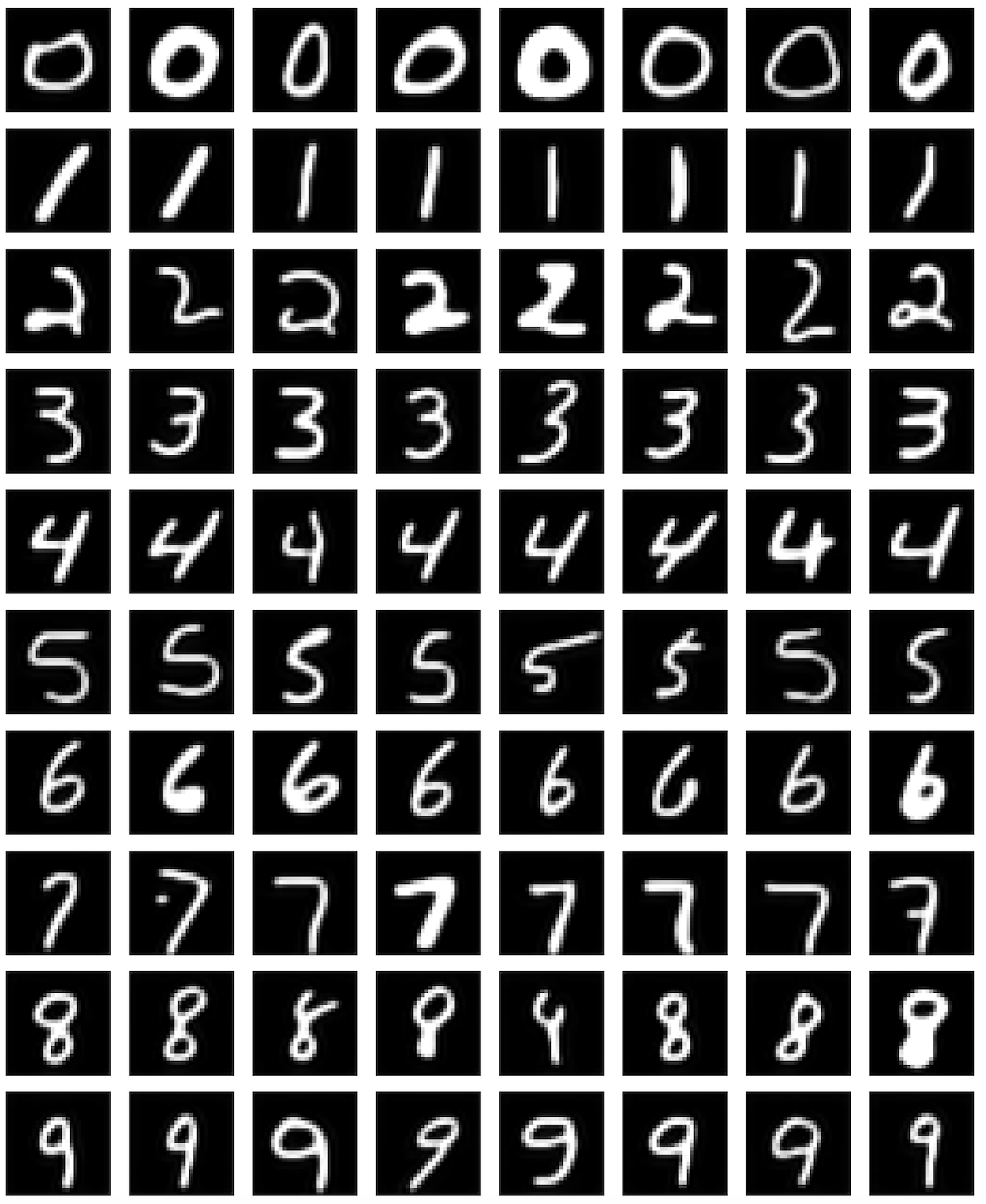}
    \caption{Two-step generated results}
    \label{fig:map_mnist_2step}
  \end{subfigure}
  \vspace{1em}
  \caption{\textbf{Generated MNIST digits.} From left: one-step and
  two-step samples from the learned transport map $T$.}
  \label{fig:combined_mnist_results}
\end{figure}

We approximate $T$ with a standard 23-million-parameter diffusion
U-Net after removing its time conditioning.
Figure~\ref{fig:combined_mnist_results} shows class-conditional samples
from the learned map. Repeated application removes some visual
artifacts and sharpens the digit boundaries.

\subsection[Latent ImageNet 256 x 256 generation]{Latent ImageNet $256\times256$ generation} \label{app:map-expt}
We use the SiT architecture \citep{ma2024sitexploringflowdiffusionbased}
without time embeddings, operating entirely in the SD-VAE latent space.
Initial ablations use a 131M-parameter B/2 model; the larger experiment
uses a 637M-parameter XL/2 model. Layer counts, hidden dimensions, and
attention heads follow the standard SiT configurations.

\paragraph{Optimization and stabilization}
We train both models with the Muon optimizer and a constant learning
rate of $10^{-3}$. We use the following stabilization techniques:

\begin{itemize}
\item \textbf{Adaptive weighting.} Following MeanFlow, we multiply the
transport loss by $\mathrm{sg}(w)$, where
$\mathcal L=\lVert\Delta\rVert_2^2$ is the regression error and
$\mathrm{sg}$ denotes stop-gradient. We set
\begin{equation}
    w = \frac{1}{(\lVert\Delta\rVert_2^2 + c)^p}.
\end{equation}
We use $p=1$ and $c=0.01$. This down-weights samples with large
regression errors during early training. The boundary term retains an
unweighted squared loss; applying the adaptive weight to it produced
blurrier samples.

\item \textbf{Gradient balancing.} Following the adaptive weighting
used in VQGAN, we balance the boundary and transport terms using the
ratio of their gradient norms with respect to the last-layer parameters
$L$:
\begin{equation}
    \lambda
    =\frac{\lVert\nabla_L\mathcal L_{\mathrm{boundary}}\rVert}
    {\lVert\nabla_L\mathcal L_{\mathrm{transport}}\rVert}.
\end{equation}
\item \textbf{ImageNet-only noisy boundary anchor.} In the direct-map
ImageNet experiments, but not in the 2D or MNIST experiments, we
augment the exact boundary loss in \eqref{eq:Tloss} by
\begin{equation}
\label{eq:noisy_anchor}
\mathcal L_{\mathrm{anchor}}(T_\theta)
=\E_{s,x_0,x_1}\!\left[
\mathbf 1_{\{s>0.8\}}\,|T_\theta(I_s)-x_1|^2
\right],
\end{equation}
using the same uniformly sampled $s$ as in the transport term. This
provides a stronger denoising signal near the data endpoint and was
useful empirically. It is an ImageNet stabilization heuristic, not part
of Theorem~\ref{thm:map_objective}. Because $I_s$ is not the autonomous
state $X_t$, a finite anchoring window need not preserve the exact
terminal map and can introduce window-dependent bias.

\end{itemize}

\paragraph{Training configurations and hyperparameters}
We train the BTM B/2 ablation model for 80 epochs and the full BTM
XL/2 model for 1280 epochs. Table~\ref{tab:hyperparameters} summarizes
the architectures and training hyperparameters.

\paragraph{Ablation and scaling results}
Table~\ref{tab:baseline} compares the 80-epoch BTM B/2 model with
MeanFlow B/2 without classifier-free guidance. Table~\ref{tab:scaling}
shows the effect of increasing the architecture to XL/2.

\paragraph{Iterated composition}
We also apply the learned map repeatedly, without re-noising, time
conditioning, or an inference schedule. As shown in
Table~\ref{tab:iterated}, additional evaluations modestly improve FID.
\begin{table}[h]
\centering

\begin{subtable}[t]{0.30\textwidth}
\centering
\begin{tabularx}{\linewidth}{@{}Xr@{}}
\toprule
\textbf{Method} & \textbf{FID} \\
\midrule
MeanFlow B/2 & 61.06 \\
BTM B/2      & 62.14 \\
\bottomrule
\end{tabularx}
\caption{80-epoch comparison without CFG.}
\label{tab:baseline}
\end{subtable}
\hfill
\begin{subtable}[t]{0.30\textwidth}
\centering
\begin{tabularx}{\linewidth}{@{}Xr@{}}
\toprule
\textbf{Method} & \textbf{FID} \\
\midrule
BTM B/2 & 62.14 \\
BTM XL/2  & 47.58 \\
\bottomrule
\end{tabularx}
\caption{Effect of model scaling at 80 epochs.}
\label{tab:scaling}
\end{subtable}
\hfill
\begin{subtable}[t]{0.30\textwidth}
\centering
\setlength{\tabcolsep}{3pt}
\begin{tabularx}{\linewidth}{@{}Xccc@{}}
\toprule
\textbf{NFE} & 1 & 2 & 5 \\
\midrule
\textbf{FID} & 17.58 & 17.04 & 16.53 \\
\bottomrule
\end{tabularx}
\caption{Iterated application of the BTM direct map.}
\label{tab:iterated}
\end{subtable}

\end{table}

\begin{table}[h]
\centering
\begin{threeparttable}
\begin{tabular}{lcc}
\toprule
\textbf{Configuration} & \textbf{BTM B/2 (Ablation)} & \textbf{BTM XL/2 (Full Scale)} \\
\midrule
\multicolumn{3}{l}{\textit{Generator Architecture}} \\
Architecture & SiT-B & SiT-XL \\
Latent Space & SD-VAE & SD-VAE \\
Input size & $32 \times 32 \times 4$ & $32 \times 32 \times 4$ \\
Patch size & $2 \times 2$ & $2 \times 2$ \\
Hidden size $d$ & 768 & 1152 \\
Layers $N$ & 12 & 28 \\
Attention Heads & 12 & 16 \\
\midrule
\multicolumn{3}{l}{\textit{Optimizer Settings}} \\
Optimizer & Muon & Muon \\
Learning Rate & $1 \times 10^{-3}$ & $1 \times 10^{-3}$ \\
Training Epochs & 80 & 1280 \\
\midrule
\multicolumn{3}{l}{\textit{Loss Stabilization}} \\
Adaptive Weight $p$ & 1 & 1 \\
Adaptive Weight $c$ & 0.01 & 0.01 \\
Boundary Loss & Plain $L_2$ + \eqref{eq:noisy_anchor} & Plain $L_2$ + \eqref{eq:noisy_anchor} \\
Gradient Balancing & Yes (last layer) & Yes (last layer) \\
\bottomrule
\end{tabular}
\caption{Hyperparameter configurations for BTM models. Architectural
dimensions match the standard SiT configurations.}
\label{tab:hyperparameters}
\end{threeparttable}
\end{table}

\end{document}